\documentclass{article}

\usepackage{arxiv}

\usepackage[utf8]{inputenc} 
\usepackage[T1]{fontenc}    
\usepackage{hyperref}       
\usepackage{url}            
\usepackage{booktabs}       
\usepackage{amsfonts}       
\usepackage{nicefrac}       
\usepackage{microtype}      
\usepackage{lipsum}		
\usepackage{graphicx}
\usepackage{natbib}
\usepackage{doi}
\usepackage{amsmath}

\title{A Novel Denoising Technique and Deep Learning Based Hybrid Wind Speed Forecasting Model for Variable Terrain Conditions}

\author{ Sourav Malakar\\
	Department of AKCSIT\\
	University of Calcutta\\
	West Bengal, India \\
	\texttt{sourav.xaviers@gmail.com} \\
	\And
	Saptarsi Goswami\\
	Bangabasi Morning College\\
	University of Calcutta\\
	West Bengal, India \\
 \And
	Amlan Chakrabarti\\
	Department of AKCSIT\\
	University of Calcutta\\
	West Bengal, India \\
 \And
	Bhaswati Ganguli\\
	Department of Statistics\\
	University of Calcutta\\
	West Bengal, India \\
}

\renewcommand{\shorttitle}{\textit{arXiv} Template}

\hypersetup{
pdftitle={A template for the arxiv style},
pdfsubject={q-bio.NC, q-bio.QM},
pdfauthor={David S.~Hippocampus, Elias D.~Striatum},
pdfkeywords={First keyword, Second keyword, More},
}

\begin{document}
\maketitle

\begin{abstract}
	Wind flow can be highly unpredictable and can suffer substantial fluctuations in speed and direction due to the shape and height of hills, mountains, and valleys, making accurate wind speed (WS) forecasting essential in complex terrain. This paper presents a novel and adaptive model for short-term forecasting of WS. The paper's key contributions are as follows: (a) The Partial Auto Correlation Function (PACF) is utilised to minimise the dimension of the set of Intrinsic Mode Functions (IMF), hence reducing training time; (b) The sample entropy (SampEn) was used to calculate the complexity of the reduced set of IMFs. The proposed technique is adaptive since a specific Deep Learning (DL) model-feature combination was chosen based on complexity; (c) A novel bidirectional feature-LSTM framework for complicated IMFs has been suggested, resulting in improved forecasting accuracy; (d) The proposed model shows superior forecasting performance compared to the persistence, hybrid, Ensemble empirical mode decomposition (EEMD), and Variational Mode Decomposition (VMD)-based deep learning models. It has achieved the lowest variance in terms of forecasting accuracy between simple and complex terrain conditions 0.70\%. Dimension reduction of IMF's and complexity-based model-feature selection helps reduce the training time by 68.77\% and improve forecasting quality by 58.58\% on average.
\end{abstract}

\keywords{Wind Forecasting \and Wind Speed Prediction \and LSTM \and EEMD}

\section{Introduction}
Wind power (WP) is one of the most popular sources of renewable energy (RE)  \cite{saidur2011environmental,ahmadi2021current,amaral2017offshore} and India has the fourth largest wind capacity in the world \cite{khan2020present}. However, rapid temporal changes cause difficulties in real-time control of WP systems  \cite{shi2013hybrid,viviescas2019contribution,abedinia2020improved,tian2020short,quan2019survey,zheng2023stochastic}. Random variation and limited predictability of WS are significant contributors to fluctuations in the safety of large-scale grid-integrated WP systems \cite{liang2022ultra, Wang2018Wind,5224014, arora2022probabilistic, yan2021frequency}. A solution can be provided by developing accurate WP forecasts \cite{yan2019advanced}. \par

  WS/WP forecasting models are classified as very short-term (a few seconds to 30 minutes in the future) \cite{shi2013hybrid}, short-term (30 minutes to 6 hours in the future), medium-term (6 hours to 1 day in the future), and long-term (more than 1 day ahead) \cite{wang2021review}. Short-term forecasting is especially useful for efficient dispatch planning and smart load-shedding decisions \cite{wang2021review}. \par

The surface of the land is referred to as terrain, often described in terms of terrain features' height, slope, and orientation. It has a chance to influence weather and climate patterns across a wide area. In complex terrain, the wind flow can be highly variable and can experience significant changes in speed and direction due to the shape and height of hills, mountains, and valleys \cite{sharples2010wind}. The surface roughness of the terrain can also have an impact on wind patterns, with rougher surfaces causing greater turbulence and wind shear \cite{tian2015terrain}. Turbulence and wind shear are caused by the interaction between the wind and the terrain \cite{tian2015terrain}. Hence, the wind flow can be highly turbulent and can experience significant wind shear change due to irregularities in the terrain. This turbulence and wind shear can cause sudden changes in WS and direction, making it difficult to accurately predict WP output. Wind turbines create wakes, regions of lower WS and higher turbulence downstream of the turbine. In complex terrain, the wakes can be highly variable and can interact with the terrain in unpredictable ways, leading to significant reductions in wind power output downstream of the turbine \cite{farrell2021design}. Addressing these challenges requires a combination of advanced modelling techniques and analysis, and innovative forecasting methods. Some potential solutions include using Machine learning (ML) algorithms to improve WP forecasting and using high-resolution models to better capture the effects of complex terrain on wind patterns.
To examine the influence of complex terrain on forecasting ability, recent state-of-the-art models \cite{huang2018wind}, \cite{shen2022wind}, and \cite{saxena2021offshore} have been implemented. The data from three stations located in simple terrain and the other two in complex have been used in this experiment. Forecasting error computed using the Normalised Root Mean Squared (nRMSE) for each station-terrain combination. The absolute forecasting deviation between simple and complex terrain was calculated using equation \ref{eq:forecasting-deviation}. The closer the FD (\%) is to zero, the more similar the forecast accuracy between the two terrains.

\begin{equation}
\resizebox{\linewidth}{!}{%
$\text{FD (\%)} = \big|\frac{(1-\frac{\sum_{i=1}^{n}\mathrm{nRMSE}_{i}^{\text{complex}}}{n}) - (1-\frac{\sum_{i=1}^{m}\mathrm{nRMSE}_{i}^{\text{simple}}}{m})}{(1-\frac{\sum_{i=1}^{m}\mathrm{nRMSE}_{i}^{\text{simple}}}{m})} \times 100$
}\big|
\label{eq:forecasting-deviation}
\end{equation}


The forecasting performance of \cite{huang2018wind}, \cite{shen2022wind}, and \cite{saxena2021offshore} is depicted in Table \ref{tab:my_label1}. It has been observed that none of the approaches has shown uniform forecasting performance in different terrain conditions. Hence, we are encouraged to design a more accurate forecasting model with consistent performance under varied terrain conditions.
\begin{table}[!t]
\centering
\caption{Deviation of forecasting accuracy between simple and complex terrain in terms of nRMSE}
\resizebox{0.9\columnwidth}{!}{%
 \begin{tabular}{|c|c|c|c|}
\hline
\textbf{Model} &\textbf{nRMSE (simple)}&\textbf{nRMSE (complex)}&\textbf{Deviation (\%)}\\

\hline
\cite{huang2018wind}& 0.1222&0.2014&9.01\\
\cite{shen2022wind} & 0.3661&0.3537&1.96\\
\cite{saxena2021offshore}& 0.1211&0.1478&3.03\\

\hline
\end{tabular}%
}
\label{tab:my_label1}
\end{table}


ML models have been widely used for short-term WS forecasting. Common ML models include Support Vector Machine (SVM) \cite{zhou2011fine}, Kalman filters \cite{zuluaga2015short}, Gaussian Regression Process (GRP) \cite{zhang2022novel}, Extreme Learning Machine (ELM) \cite{hua2022integrated}, multi-layer perceptron (MLP) \cite{samadianfard2020wind}, and  Artificial Neural Networks (ANN) \cite{navas2020artificial}. WS is non-stationary and non-linear \cite{lv2011short, li2022innovative,hu2021new} and is considered one of the most complex meteorological parameters to forecast. As deep learning (DL) models can handle complex nonlinear structures, they have shown lower forecasting errors than many traditional ML models \cite{abdulla2022design}. Some commonly used DL models are Long Short-Term Memory (LSTM) \cite{memarzadeh2020new}, Convolution Neural Networks (CNN) \cite{wang2020short, yan2021frequency}, Bi-directional LSTM (Bi-LSTM) \cite{saxena2021application}, Gated Recurrent Units (GRU) \cite{syu2020ultra}, Bi-directional GRUs (Bi-GRU) \cite{xu2022multi}, etc.  \cite{shen2022wind} proposes a hybrid framework based on CNN and LSTM for multi-step WS forecasting. The proposed CNN-LSTM  achieves the lowest forecasting error as compared to the Recurrent Neural Network (RNN), LSTM, and CNN.

 Signal processing techniques have been commonly used to pre-process WS data. Typically, these techniques can be classified into wavelet-based methods, mode decomposition-based techniques and single spectrum analysis (SSA) \cite{hassani2007singular}-based models. Mode decomposition-based methods include VMD \cite{dragomiretskiy2013variational,9384295,li2019short,fu2021multi,wang2019deep}, Empirical Mode Decomposition (EMD) \cite{rilling2003empirical,ren2014novel}, EEMD \cite{wu2009ensemble}, etc. \cite{manjula2012comparison} mentions that wavelet-based techniques are localized in time and frequency producing coefficients at exceptional scales. Nevertheless, wavelet transforms can be unreliable in the presence of noise. On the other hand, mode decomposition-based methods are useful for non-stationary and non-linear time series \cite{wu2009ensemble}. \cite{huang2018wind} develops a short-term WS prediction model using LSTM, Gaussian process regression (GPR), and EEMD. The model outperforms single LSTM and single GPR models. \cite{saxena2021offshore} proposes a short-term forecasting model for offshore wind forecasting based on EEMD and Bi-LSTM. This EEMD-(Bi-LSTM) model is shown to be more accurate as compared to EEMD-LSTM, EEMD-CNN, EEMD-CNNLSTM, CNN, and CNN-LSTM. In paper \cite{li2022novel}, the authors have developed a particle swarm optimization (PSO)-VMD-Bi-LSTM based wind forecasting model for accurate prediction of short-term typhoon WS. In the presence of high levels of noise, EEMD is known to perform better due to its adaptive nature, While VMD may have trouble separating modes. EEMD does not require any prior knowledge about the frequency content of the signal, while VMD assumes smooth variations in the signal. Due to the chance of highly stochastic WS in complicated terrain, EEMD has been used in this study as one of the pre-processing steps to separate the WS into IMF components.

According to \cite{chen2021short}, the scale and non-linearity of WS are on the rise as it is being measured for longer periods. As a result, the number of sub-components (IMFs) will increase if EEMD is used. The following bottlenecks have been reported in the literature: 

\begin{itemize}
 \item If DL algorithms are used to predict all IMFs produced by EEMD, the total number of trainable parameters will increase significantly \cite{duan2022combined}.
  
    \item Use of genetic algorithm (GA) based solutions is expensive and will not ensure optimal IMF selection. 
    
    \item Prediction of all IMFs via the same deep learning model will not lead to optimal forecasting performance.

\item Prior research provides no information on variations in model performance for different terrains and seasons.

\end{itemize}

An adaptive short-term WS forecasting model has been developed in this paper. As a pre-processing step, EEMD has been applied to the raw WS data to remove noise by discarding high-frequency IMFs. The IMF count has been reduced by calculating optimal lag using PACF \cite{lee2013short}, which further helps reduce training time. SampEn has been used to select an appropriate DL model-feature combination for each IMF. 

Our contributions can be listed as follows: 

\begin{itemize}

  \item A novel short-term adaptive forecasting model for complex terrain conditions has been proposed.
    
    \item The training time is also reduced using PACF-based optimal lag computation in addition to the IMF count reduction.
 
    \item Both simple and complex IMFs are not modelled using a single model as \cite{saxena2021offshore, huang2018wind}. The proposed technique is adaptive since a particular DL model-feature combination has been chosen depending on complexity.

    \item A unique bidirectional feature-LSTM framework has been proposed for complex IMFs, leading to better forecasting accuracy. 
    
    \item The proposed model has been validated for five wind stations in plain and complex terrains for summer, winter, and rainy seasons.
    
    \item The forecasting performance of the proposed model has been shown for both single and multi-step ahead forecasting. Also, forecasting superiority has been compared with the Persistence \cite{luo2018short}, unidirectional LSTM, and both EEMD and VMD-based hybrid deep learning models.
\end{itemize}

The rest of the paper is organized as follows. First, SampEn has been thoroughly discussed in Section \ref{bg}. Next, the working principle of EEMD has been discussed. Subsequently, we have provided a brief outline of the LSTM architecture. In Section \ref{mm}, we have discussed the materials and methods employed in setting up the empirical study. Moreover, different phases of the proposed model architecture are discussed. In Section \ref{res}, the forecasting performance of the models is critically analyzed and discussed, and finally, in Section \ref{con}, the concluding remarks are presented.
\section{Background}
\label{bg}
\subsection{Sample entropy (SampEn)}
\label{sampen}
Sample entropy (SampEn) is a complexity measure of time-series signals \cite{song2010new}. The SampEn is more resistant to noise \cite{richman2000physiological}. It is also robust to outliers such as spikes. SampEn is an interesting tool for the nonlinear analysis of time-series signals because of these properties.

Let's assume we have a time-series data of length \textit{N} = \{$X_1, X_2, X_3, ..., X_N$\} with a constant time interval $\tau$. A template vector of length \textit{m} and the distance measure are defined as $X_m(i) = \{X_i, X_{i+1}, X_{i+2}, ..., X_{i+m-1}\}$, and 
$d[X_m(i), X_m(j)], i \neq j$. The SampEn is defined in the equation \ref{samp}.

\begin{equation}
    SampEn(m, r, \tau) = -ln\frac{A}{B}
    \label{samp}
\end{equation}

Where, 
\par
A = Number of template vector pairs having $d[X_{m+1}(i), X_{m+1}(j)]<r$.
\par
B = Number of template vector pairs having $d[X_{m}(i), X_{m}(j)]<r$.
\par
SampEn will always be either a zero or a positive number. A lower SampEn number suggests less noise. 

\subsection{Ensemble Empirical Mode Decomposition (EEMD)}
\label{EEMD}   
EMD is a noise-removal method that works well for non-stationary and linear time series. It filters out oscillation modes with different frequency bands known as IMFs of the same length as the original signal in the time domain \cite{bagherzadeh2021analysis}. However, EMD suffers from a mode mixing problem \cite{4566821}. EEMD \cite{wu2009ensemble} has been proposed as a solution, and implementation involves the following steps:

\begin{itemize}
\item \textbf{Step 1:} Add white noise to the original time series.
    \item \textbf{Step 2:} Use the EMD method to decompose the noisy signal into IMFs.
    \item \textbf{Step 3:} Repeat steps 1 and 2, n times by adding a different white noise generation to obtain n sets of IMF and residual components.
    \item \textbf{Step 4:} Compute ensembles of n sets of IMF components. The added white noise  cancels out, reducing the chance of mode mixing.   
\end{itemize}

\subsection{Long Short-Term Memory (LSTM)}
\label{LSTM}

LSTMs are a special type of RNN  designed to learn both long- and short-term dependencies~\cite{schmidhuber1997long}. Compared to a traditional neural network, LSTM units encompass a `memory cell' that can retain and maintain information for long periods \cite{sharma2021sequential}. A group of gates is used to customize the hidden states, input, forget, and output. The functionality of each gate is summarized as follows.

\begin{itemize}
 
    \item \textbf{{Forget} gate $(f_t) = \sigma(w_f[h_{t-1}, x_t]+b_f)${:} } On the basis of certain conditions such as $x_t$, $h_{t-1}$, and~a sigmoid layer, a  forget gate produces either 0 or 1. If~ 1, memory information is preserved; otherwise, it is discarded.
    \item \textbf{{Input} gate~$(i_t) = \sigma(w_i[h_{t-1}, x_t]+b_i)$:}   helps in deciding which values from the input are used for the current memory state.   
        \item \textbf {{Cell} state~$(c_t) = tanh(w_c[h_{t-1}, x_t]+b_c)$:}  new cell state $c_t$ is the summation of $c_{t-1}*f_t$, and~$\widehat{c_t }*i_t$. $c_{t-1}*f_t$ decides the fraction of the old cell state that  is discarded, and the amount of new information that  is added is decided through $\widehat{c_t }*i_t$. 
    \item \textbf{{Output} gate~$(o_t) = \sigma(w_o[h_{t-1}, x_t]+b_o)$:}   decides what to output  on the basis of the current input and previous hidden state.
    \item \textbf{{Hidden} state~$(h_t) = o_t*tanh(c_t)$:}  current hidden state is computed by multiplying the output gate by the current cell state using the tanh function. 
\end{itemize}
Here, $w_f$, $w_i$, $w_c$ and $w_o$ are  weight matrices. $b_f$, $b_i$, $b_c$ and $b_o$ are the biases for individual gates. $\sigma$ indicates a sigmoid activation function. * stands for element-wise multiplication and + implies element-wise addition. 

\section{Materials and methods}
\label{mm}
In this section, the dataset, the proposed model architecture, and forecasting performance evaluation criteria have been discussed in detail. 

\subsection{Motivating dataset }
\label{ED}
WS [m/s] data of 1-hour resolution has been collected from five wind stations of India from  Modern-Era Retrospective analysis for Research and Applications, Version 2 (MERRA-2) \cite{gelaro2017modern}, NASA for the period, 01/01/2015 - 31/12/2021. Table \ref{tab:my_label65} summarises the information by location and terrain.

\begin{table}[!t]
\centering
\caption{Summary of WS[m/s] distribution by station}
\resizebox{1.0\columnwidth}{!}{%
 \begin{tabular}{|c|c|c|c|c|c|c|}
\hline
  
\textbf{Lat. \& Lon.} &\textbf{Station} &\textbf{Terrain}&\textbf{Minimum}&\textbf{Maximum}&\textbf{Standard deviation}&\textbf{Mean}\\

\hline
 22.27 and 72.20&Motibaru& Plain&0.02000&22.27000&2.26878&5.08357\\
  8.96 and 77.71&Kayathar& Plain&0.02000&16.56000&2.25249&5.03836\\
   
  8.28 and 77.59&Kavalkinaru& Plain&0.02000&22.64000&2.36418&5.49630\\
  17.37 and 73.48&Jagmin& Complex&0.05000&18.66000&2.46950&5.09041\\
    18.18 and 76.04&Gadhpati& Complex&0.01000&15.87000&2.46999&5.38728\\
\hline
\end{tabular}%
}

    \label{tab:my_label65}
\end{table}








\subsection{EEMD decomposition, IMF reduction and classification }
\label{eemd}

\begin{figure}[]

\renewcommand{\arraystretch}{0.5}
\setlength\tabcolsep{-10.5pt}
\centering
\begin{tabular}{c}

\includegraphics[width=8cm,height=8cm]{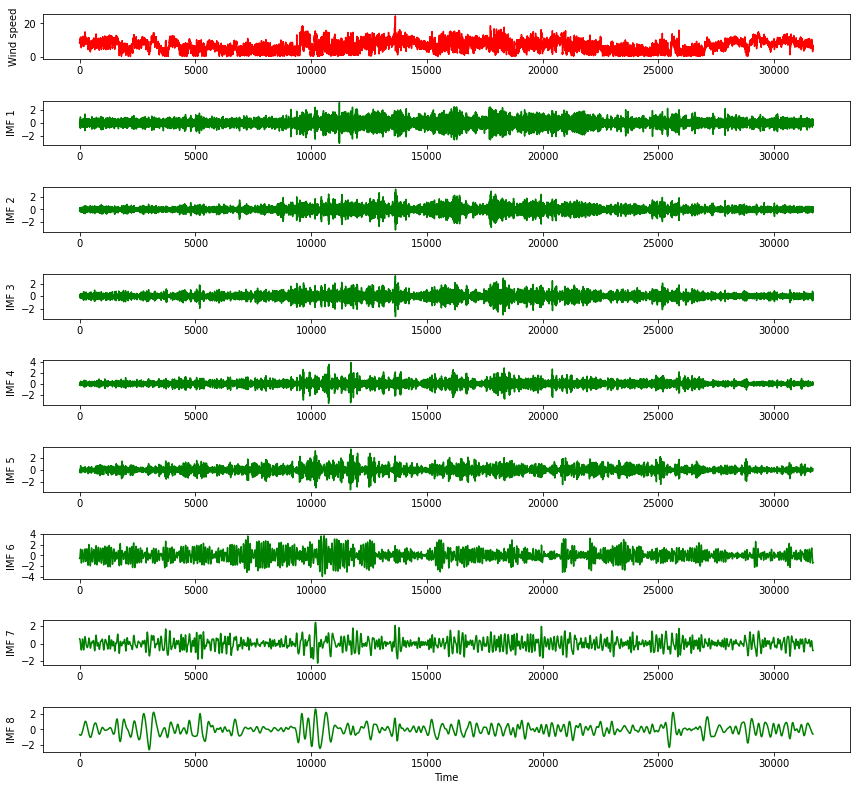} \\[1pt]\includegraphics[width=8cm,height=8cm]{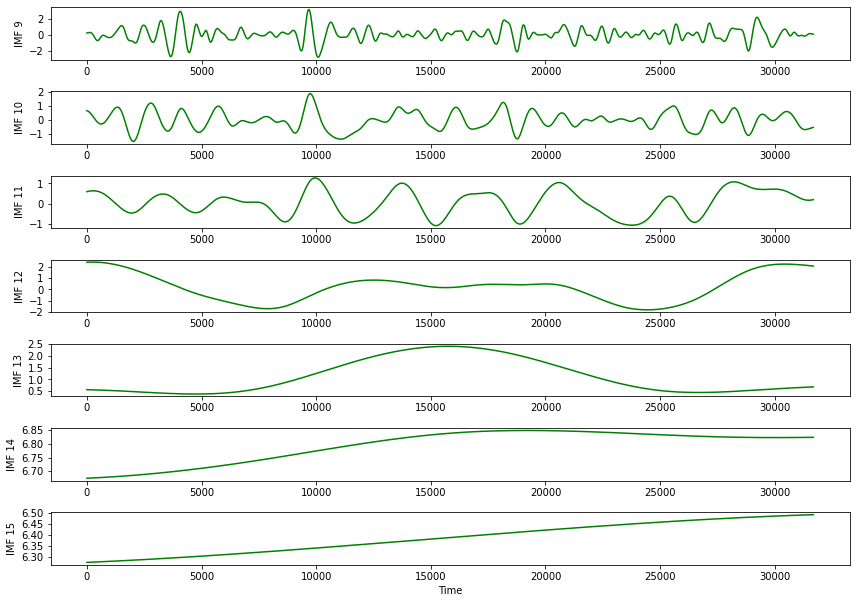} \\[1pt]


(a) EEMD decomposition\\
\includegraphics[width=8cm,height=6cm]{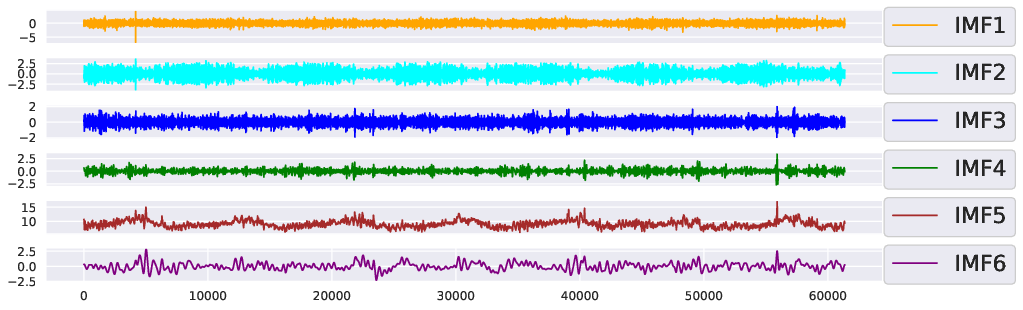} \\[1pt]
(b) Reduced set of IMFs\\
\end{tabular}
 \caption{ EEMD decomposition and reduction for Gadhpati }
 
\label{data}
\end{figure}

EEMD has been applied to the WS data to decompose it into several stationary sub-signals, termed IMFs and a residual. For this decomposition, we have used the default noise and ensemble numbers as 0.05 and 100. Figure \ref{data} (a) shows the decomposition result for Gadhpati station. As all the IMFs are stationary, PACF has been applied to each IMF to obtain optimal lag values. Next, IMFs having identical lags are combined to obtain a new signal. The above technique reduces the total number of IMFs, shown in Figure \ref{data} (b). The SampEn \cite{richman2004sample} metric is used to calculate individual IMF complexity and a deep learning model-feature combination is chosen based on the complexity score.

\begin{figure}[t!]
\centering
\includegraphics[width=65mm]{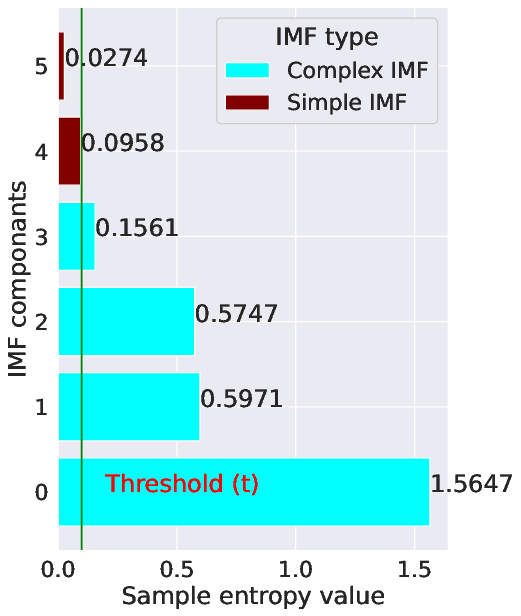}
\caption{Complex and simple IMF identification}
\label{imf1}
\end{figure}

 Figure \ref{imf1} shows the complexity scores of all the IMFs for Gadhpati. A predefined threshold of 0.1 \textit{t}, has been taken to classify the IMFs into two groups. If the complexity score of a particular IMF is higher than 0.1, then it is classified as complex and otherwise marked as simple. 

\subsection{Data preprocessing for deep learning }
\label{dp}

\begin{figure}[t!]

\centering

\includegraphics[width=90mm]{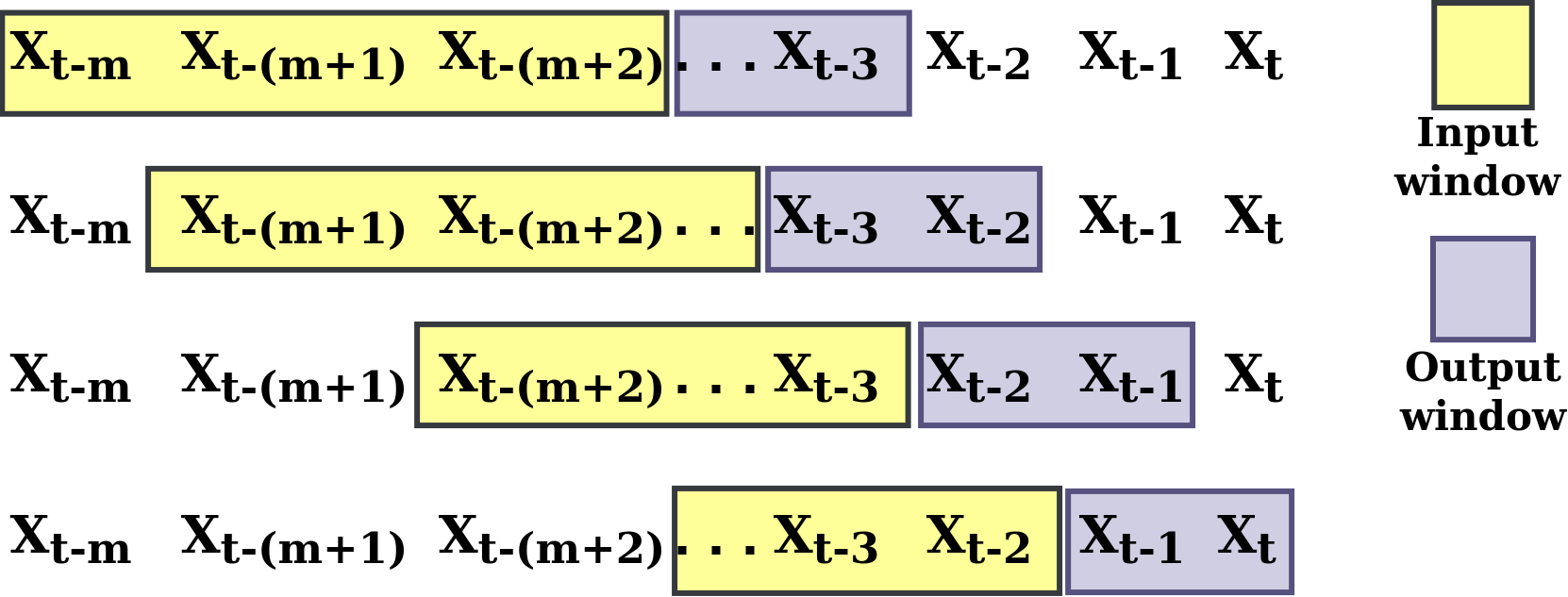}

\caption{The sliding window approach}
\label{sliding}
\end{figure}

A sliding window approach has been used to prepare the time-series data for deep learning and is shown in Figure \ref{sliding}. Suitable window sizes, \textit{m}, and \textit{n} have been chosen for the segmentation, where \textit{m} is the size of the input window and \textit{n} is the size of the output window. The input window covers {\emph{m}} past observations, \{$X_1$, $X_2$, $X_3$, ..., $X_m$\}, and is used to predict the next \textit{n} observations, \{$X_{m+1}$, ..., $X_{m+n}$\}. Next, the input window is shifted one position to the right as \{$X_2$, $X_3$,  $X_4$, ..., $X_{m+1}$\}, and \{$X_{m+2}$, ..., $X_{m+n+1}$\}. This process continues until no data points are left.

Data normalization is essential for stable and improved forecasting performance of deep learning models. Min-Max normalization \cite{patro2015normalization} has been applied to the data to scale it to lie in [0, 1].

\subsection{Proposed forecasting framework }
\label{ff}
\begin{figure*}[h!]

\centering

\includegraphics[width=160mm]{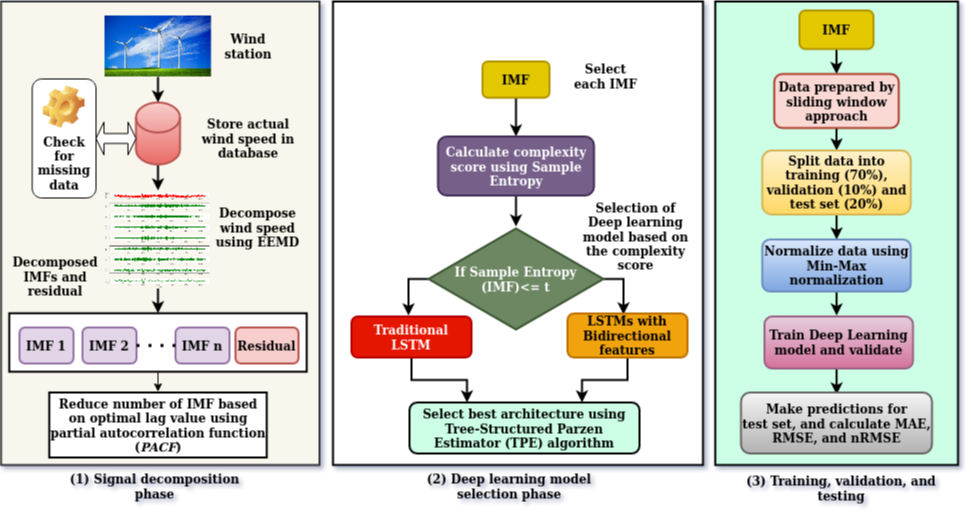}

\caption{The proposed model architecture}
\label{var}
\end{figure*}

The workflow of the proposed model has been presented in Figure \ref{var}. It consists of the following steps.

\textbf{Training phase:}

\begin{itemize}
 \item EEMD has been applied to decompose raw WS data into k IMFs given by $IMF_1$, $IMF_2$, $IMF_3$, ..., $IMF_k$, and a residual, denoted by $R_k$.
    
    \item The optimal lag is computed for each IMF using the PACF. By superimposing IMFs with the same lag, the total number of IMFs has decreased. 

\item SampEn is used to calculate the complexity of each new IMF. IMFs are classified as complex or simple by comparison with a threshold which we set at 0.1.  

\item  For simple IMFs presented in Figure \ref{blstm} (a), to predict an observation at time step \textit{t} of $Day_0$ denoted by $X_t$, we use  previous \textit{m} observations, $Day_0$ [$X_{t-1}$, $X_{t-2}$, $X_{t-3}$, ..., $X_{t-m}$] as also the observations for the same time frame from the previous day, $Day_{-1}$[$X_{t-1}$, $X_{t-2}$, $X_{t-3}$, ..., $X_{t-m}$]. Complex IMFs have been modelled using two parallel LSTM networks presented in Figure \ref{blstm} (b). In case of the backward LSTM, \textit{n} future observations, denoted by $Day_{-1}$[$X_{t+1}$, $X_{t+2}$, $X_{t+3}$, ..., $X_{t+n}$] have been augmented from the previous day.

\begin{figure}[t!]

\centering

\includegraphics[width=90mm]{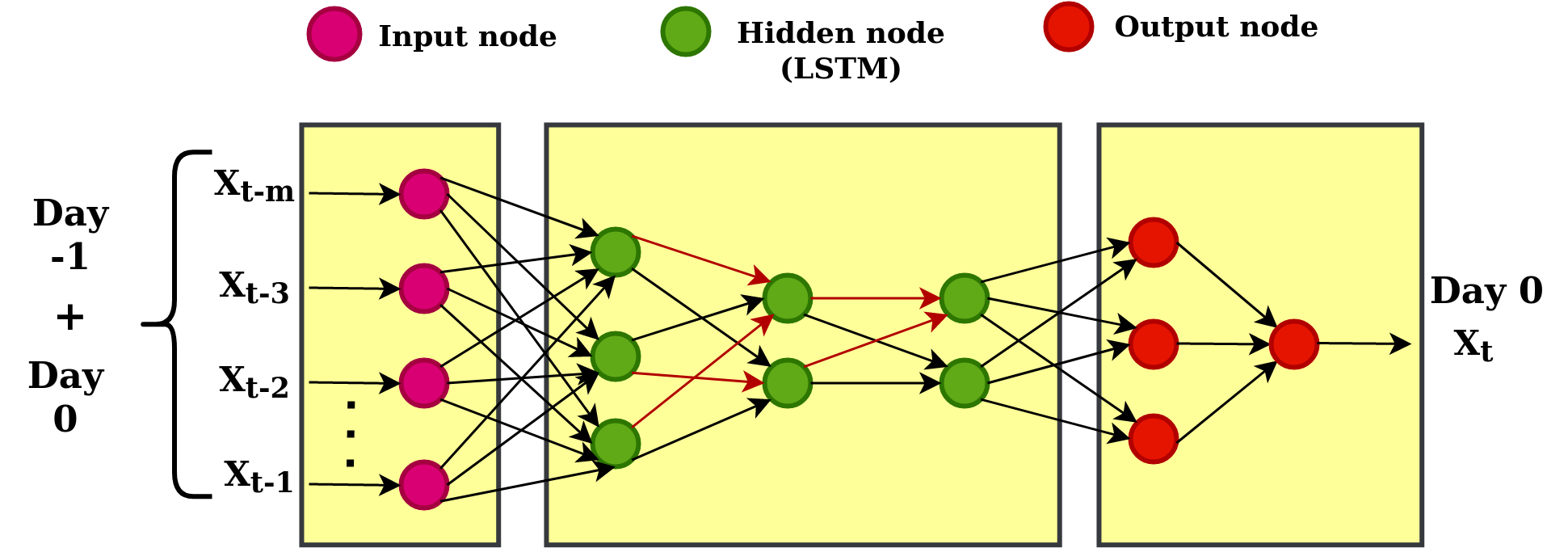}\\
(a) Architecture of traditional LSTM-based neural network\\
\includegraphics[width=90mm]{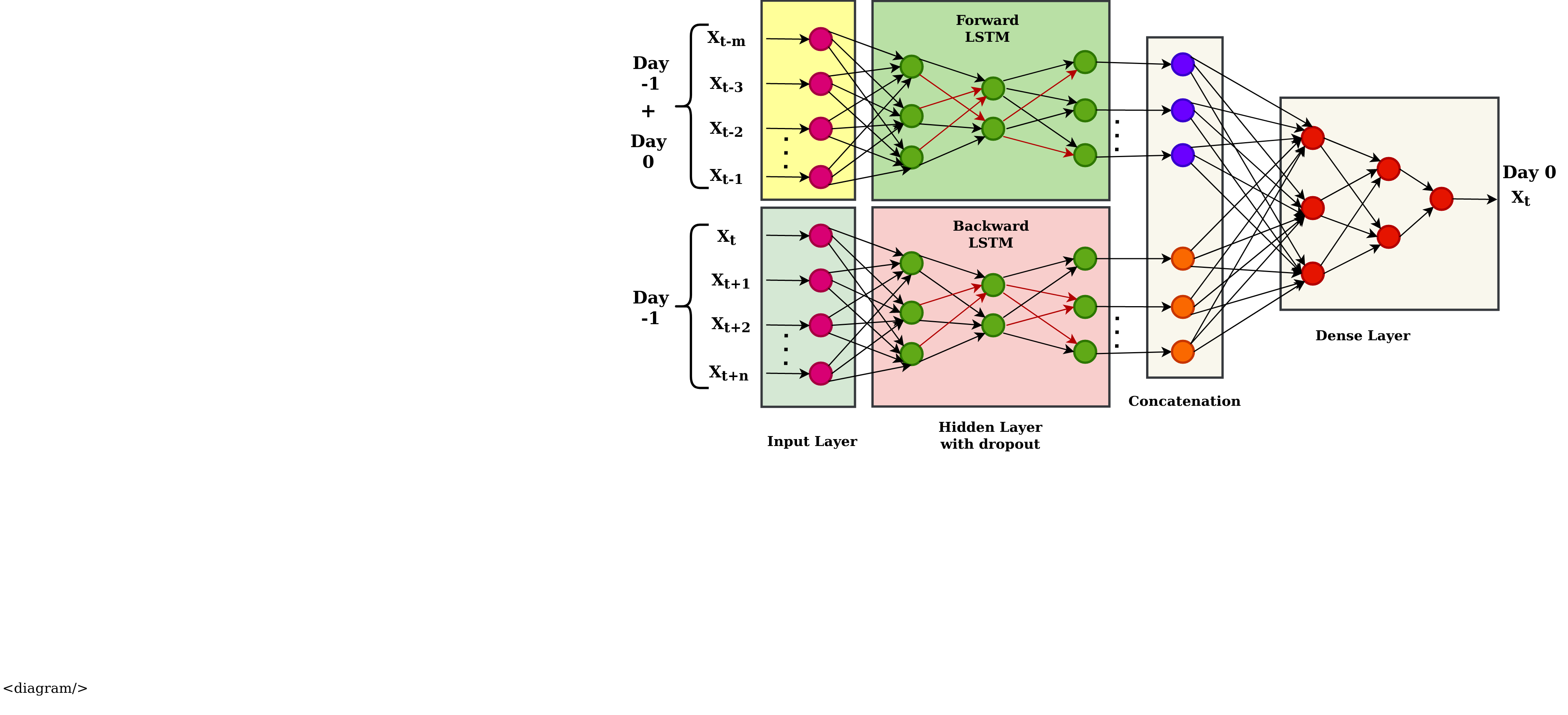}\\
(b) Architecture of bidirectional feature and LSTM-based neural network

\caption{Neural network part of the proposed model }
\label{blstm}
\end{figure}
    
   \item For each IMF, the sliding window method has been used to prepare the data for deep learning. Min-Max normalization is used to normalize the data. Models were trained using a split into training (70\%) and validation (10\%) data. 

\end{itemize}

\textbf{Testing phase:}

\begin{itemize}

\item Figure \ref{RTF}, based on \cite{chen2022decomposition}, illustrates the process of real-time forecasting. Every time a new instance appears, it is added to the existing time series, and the updated series is then divided into sub-sequences.  

\item The proposed model has been tested with 20\% of the data. An ensemble of predictions is used to obtain the forecast WS.

\begin{figure}[!th]

\centering

\includegraphics[width=70mm]{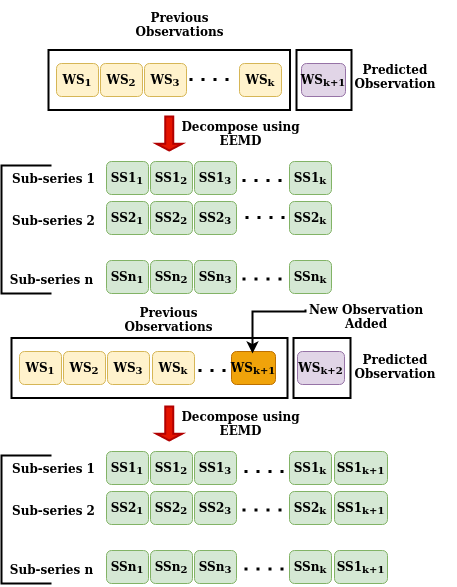}

\caption{Real time WS[m/s] decomposition}
\label{RTF}
\end{figure}
\end{itemize}

\subsection{Evaluation of forecasting model}
\label{fp}
Forecasting performance has been evaluated using three standard metrics, namely, Mean Absolute Error (MAE), Root Mean Squared Error (RMSE), and normalized Root Mean Square Error (nRMSE). 
\begin{equation}
    \mbox{MAE} = \frac{1}{N}\sum_{i=1}^{N}\left | \hat{y}_{i} - y_{i} \right |
\end{equation}
\begin{equation}
    \mbox{RMSE }= \sqrt{\frac{\sum_{i=1}^{N}( \hat{y_i}-y_i)^2}{N}}
\end{equation}
\begin{equation}
       \mbox{nRMSE} =\frac{ \sqrt{\frac{\sum_{i=1}^{N}( \hat{y_i}-y_i)^2}{N}}}{\mu}
\end{equation}
\section{Results and discussion}
\label{res}
A machine with 16 GB RAM and an Intel Core i5-8250U CPU running at 1.60 GHz was used for analysis.  Deep learning models have been implemented using Keras \cite{gulli2017deep} and TensorFlow \cite{abadi2016tensorflow} in Python. Visualization is done using Matplotlib \cite{hunter2007matplotlib}, and Seaborn \cite{bisong2019matplotlib} packages. Specific hyper-parameters (number of hidden layers, number of nodes in each layer, batch size, number of epochs, dropout and learning rates) have been optimized using the Tree-structured Parzen Estimator (TPE) \cite{bergstra2011algorithms} algorithm. Table \ref{tab:my_label107} presents optimal choices of the hyper-parameters with two additional model features. As described in subsection \ref{eemd}, PACF was used to lower the IMF count for the proposed model, which does not apply to other benchmark models. A smaller number of IMFs results in reduced training time for the model. The suggested model also included a dynamic model-feature selection method, as stated in subsection \ref{ff}. 

\begin{table}[t!]

\centering
\caption{Optimal hyper-parameter settings with some additional features of the proposed and benchmark models }
\resizebox{1.0\columnwidth}{!}{%
\begin{tabular}{|c|ccccc|}
\hline

  \textbf{Models}& \textbf{Deep Learning framework}&\textbf{Hyper-parameters}&\textbf{\shortstack{Value}} &\textbf{\shortstack{IMF Count Reduction}}&\textbf{\shortstack{Dynamic Model Selection}}\\ 
\hline
Proposed approach &LSTM&Number of hidden layers&2&&\\
 &&Nodes in hidden layer one&75&&\\
&&Nodes in hidden layer two &65&&\\
&&Learning rate &0.0001&&\\
 &&Batch size&100&&\\
 &&Epochs&50&&\\
 &&Dropout&0.2&&\\
 &&Activation&'tanh'&&\\
 &&Stateful&True&Yes&Yes\\
 &LSTMs with bidirectional features&Number of hidden layers&2&&\\
 &&Nodes in hidden layer one&25&&\\
&&Nodes in hidden layer two &25&&\\
&&Learning rate &0.001&&\\
 &&Batch size&100&&\\
 &&Epochs&25&&\\
 &&Dropout&0.02&&\\
  &&Activation&'tanh'&&\\
 &&Stateful&True&&\\
\hline
\cite{elsaraiti2021comparative}&LSTM &Learning rate &0.01&No&No\\
\hline
\cite{huang2018wind}&EEMD-LSTM &Number of hidden layers&2&&\\
 &&Nodes in hidden layer one&25&No&No\\
&&Nodes in hidden layer two &25&&\\
\hline
\cite{saxena2021offshore}&EEMD-(Bi-LSTM) &Number of hidden layers&2&&\\
 &&Nodes in hidden layer one&25&&\\
&&Nodes in hidden layer two &25&No&No\\
&&Learning rate &0.001&&\\
 &&Batch size&100&&\\
 &&Epochs&15&&\\
\hline
\cite{shen2022wind}&(1D-CNN)-LSTM &Filter 1&  (1,32)     && \\
 &&Filter 2&  (1,32) &&\\
&&Nodes in hidden layer one &25&&\\
&&Learning rate &0.001&No&No\\
 &&Batch size&100&&\\
 &&Epochs&15&&\\
 &&Activation&'ReLU'&&\\
 &&Dropout&0.2&&\\

\hline
\cite{li2022novel}&VMD-(Bi-LSTM) &Number of hidden layers&2&&\\
 &&Nodes in hidden layer one&25&&\\
&&Nodes in hidden layer two &25&No&No\\
&&Learning rate &0.001&&\\
 &&Batch size&100&&\\
 &&Epochs&15&&\\
\hline
\end{tabular}%
}

    \label{tab:my_label107}
\end{table}

\subsection{Comparison with benchmark models}
The forecasting performance of the proposed model has been compared against persistence, non-EEMD, and EEMD-based deep learning approaches as shown in Table \ref{tab:my_label59}. In the case of a plain terrain, the proposed model has shown 79.35\%, 85.06\%, 19.55\%, 19.13\%, and 73.37\% greater forecasting accuracy as compared to persistence, \cite{elsaraiti2021comparative,huang2018wind,saxena2021offshore}, and \cite{shen2022wind} in terms of nRMSE. It has outperformed others by 79.07\%, 74.95\%, 10.46\%, 3.83\%, and 73.02\% in terms of RMSE. Also, it has dominated other models by 78.81\%, 74.94\%, 11.35\%, 5.25\%, and 73.25\%. in terms of MAE. In the case of complex terrain, the proposed model has achieved 78.50\%, 85.09\%, 43.20\%, 31.33\%, and 71.25\% lower forecasting error compared to persistence, \cite{elsaraiti2021comparative}, \cite{huang2018wind}, \cite{saxena2021offshore}, and \cite{shen2022wind} in terms of nRMSE. In terms of RMSE, the proposed model achieved 78.51\%, 73.26\%, 28.24\%, 31.33\%, and 71.26\% superior forecasting quality compared to others. In terms of MAE, the proposed model has outperformed others by 78.26\%, 72.51\%, 20.12\%, 31.42\%, and 70.74\%. The above discussion suggests that the proposed model has superior forecasting performance for both terrains compared to the benchmark models. The second best performing model is EEMD-(Bi-LSTM) \cite{saxena2021offshore}.

\begin{table}[t!]
\centering
\caption{Four hours ahead forecasting accuracy }
\hspace*{-2.5cm} 
\begin{tabular}{|c|c|c|c|c|c|c|c|}
\hline
\shortstack{Wind \\stations}& \shortstack{Error \\metrics}&Persistence&\shortstack{LSTM \\\cite{elsaraiti2021comparative}}& \shortstack{EEMD-LSTM \\\cite{huang2018wind}}& \shortstack{EEMD-(BiLSTM)\\\cite{saxena2021offshore} }& \shortstack{(1D-CNN)-LSTM\\\cite{shen2022wind}}& \shortstack{Proposed \\approach}\\ 
\hline
    \multicolumn{8}{|c|}{Plain terrain} \\ 
\hline
&RMSE&0.9618&0.8067&0.2026&0.1971&0.7389&\bf0.1947\\
Kayathar&nRMSE&0.4464&0.3744&0.0940&0.0914&0.3429&\bf0.0868\\
&MAE&0.6858&0.5874&0.1466&0.1452&0.5382&\bf0.1432\\
\hline
&RMSE&0.9651&0.8300&0.2936&0.3190&0.7672&\bf0.2522\\
Kavalkinaru&nRMSE&0.4192&0.3605&0.1274&0.1384&0.3332&\bf0.1094\\
&MAE&0.6894&0.6045&0.2205&0.2347&0.5692&\bf0.1888\\
\hline
&RMSE&1.3376&1.0636&0.3475&0.3197&1.0098&\bf0.2192\\
Motibaru&nRMSE&0.5594&0.4448&0.1453&0.1337&0.4223&\bf0.0917\\
&MAE&0.9822&0.7735&0.2603&0.2386&0.7384&\bf0.1601\\
\hline
    \multicolumn{8}{|c|}{Complex terrain} \\ 
\hline
&RMSE&1.1991&0.9696&0.3361&0.3482&0.8584&\bf0.2115\\
Jagmin&nRMSE&0.4710&0.3809&0.2520&0.1369&0.3372&\bf0.0832\\
&MAE&0.8747&0.7039&0.1963&0.2610&0.6336&\bf0.1572\\
\hline
&RMSE&1.1767&0.9416&0.3698&0.3891&0.9077&\bf0.2980\\
Gadhpati&nRMSE&0.4799&0.3840&0.1508&0.1587&0.3702&\bf0.1215\\
&MAE&0.8639&0.6751&0.2765&0.2864&0.6536&\bf0.2203\\
\hline
\end{tabular}
\label{tab:my_label59}
\end{table}

\begin{figure}[!htb]

\centering

\includegraphics[width=90mm]{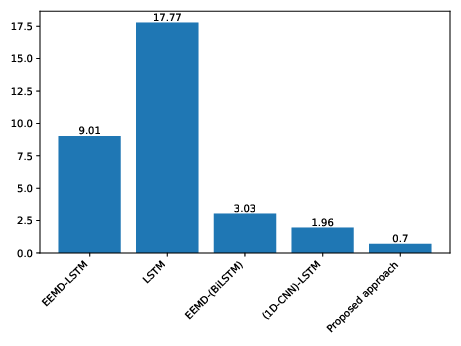}

\caption{Deviation in nRMSE between simple and complex terrain}
\label{terrain}
\end{figure}

Figure \ref{terrain} depicts the deviation in forecasting accuracy between simple and complex terrain conditions. It is computed based on \ref{eq:forecasting-deviation}. The proposed model has achieved the lowest deviation in forecasting accuracy in varying terrain conditions.

\subsection{Forecasting performance on multi-steps ahead WS prediction}

The forecasting performance of the proposed model has been compared with the benchmark models for four different choices of forecasting horizons - 1 hr, 2 hrs, 3 hrs, and 4 hrs.  In case of 1 hr ahead forecasting the proposed model has achieved 79.39\%, 74.76\%, 25.34\%, 15.50\%, and 75.68\% lower nRMSE compared to persistence, \cite{elsaraiti2021comparative}, \cite{huang2018wind}, \cite{saxena2021offshore}, and \cite{shen2022wind}. In the case of 2 hrs-ahead forecasting, the improvement is 80.72\%, 73.87\%, 23.49\%, 16.15\%, and 77.40\%. In the case of 3 hrs ahead forecasting, the proposed model has dominated others by 85.24\%, 81.53\%, 31.07\%, 21.09\%, and 79.90\%. In the case of 4 hrs ahead forecasting, the benchmark models have been outperformed by the proposed approach by 79.53\%, 78.52\%, 24.82\%, 18.24\%, and 76.42\%. 

\begin{table}[t!]

\centering
\caption{Forecasting performance of the proposed model for 1-4 hr ahead}
\resizebox{1.0\columnwidth}{!}{%
\begin{tabular}{|c|c|c|c|c|c|c|c|}
\hline
\shortstack{Wind\\stations}& \shortstack{Forecasting \\step}&Persistence&\shortstack{LSTM \\\cite{elsaraiti2021comparative}}& \shortstack{EEMD-LSTM \\\cite{huang2018wind}}& \shortstack{EEMD-(BiLSTM)\\\cite{saxena2021offshore} }& \shortstack{(1D-CNN)-LSTM\\\cite{shen2022wind}}& \shortstack{Proposed \\approach}\\ 
\hline

    \multicolumn{8}{|c|}{Plain terrain} \\ 
\hline
&step 1&0.2135&0.1884&0.0495&0.0473&0.1839&\bf0.0613\\
Kayathar&step 2&0.3713&0.3145&0.0783&0.0797&0.2955&\bf0.0645\\
&step 3&0.5991&0.4198&0.0942&0.0995&0.3820&\bf0.0904\\
&step 4&0.6039&0.5000&0.1264&0.1299&0.4509&\bf0.1185\\
\hline
&step 1&0.2026&0.1752&0.1096&0.1136&0.1870&\bf0.0988\\
Kavalkinaru&step 2&0.3486&0.2997&0.1292&0.1314&0.2869&\bf0.1092\\
&step 3&0.4684&0.4044&0.1277&0.1372&0.3701&\bf0.1101\\
&step 4&0.5664&0.4854&0.1412&0.1663&0.4358&\bf0.1188\\
\hline
&step 1&0.2566&0.1992&0.1015&0.0934&0.2060&\bf0.0508\\
Motibaru&step 2&0.3677&0.1887&0.1198&0.1074&0.3539&\bf0.0759\\
&step 3&0.6258&0.5028&0.1608&0.1431&0.4750&\bf0.0910\\
&step 4&0.7637&0.6032&0.1843&0.1754&0.5658&\bf0.1303\\
\hline

    \multicolumn{8}{|c|}{Complex terrain} \\ 
\hline
&step 1&0.2157&0.1673&0.1045&0.1103&0.1774&\bf0.0988\\
Jagmin&step 2&0.3105&0.2613&0.1309&0.1387&0.2817&\bf0.1190\\
&step 3&0.5269&0.4295&0.1385&0.1371&0.3747&\bf0.1182\\
&step 4&0.6437&0.5209&0.1503&0.1573&0.4511&\bf0.1267\\
\hline
&step 1&0.2179&0.1699&0.1153&0.1151&0.1776&\bf0.0977\\
Gadhpati&step 2&0.3912&0.3117&0.1537&0.1545&0.3045&\bf0.1228\\
&step 3&0.5361&0.4324&0.1607&0.1672&0.4137&\bf0.1270\\
&step 4&0.6582&0.5261&0.1679&0.1887&0.5029&\bf0.1353\\
\hline

\end{tabular}%
}

    \label{tab:my_label60}
\end{table}

\subsection{Season specific forecasting performance}
Table \ref{tab:my_label80} shows season-specific performance for all forecasting stations in terms of nRMSE. For stations located in plain terrain, the proposed model has achieved 70.80\%, 30.01\%, 7.65\%, and 46.16\% lower nRMSE in winter as compared to the benchmarks. In summer, the other models have been dominated by 81.23\%, 48.23\%, 23.32\%, and 79.12\%. In the rainy season, the proposed model has shown 82.46\%, 37.70\%, 15.79\%, and 79.07\% lower forecasting error in terms of nRMSE. In the case of complex terrain, the proposed model has shown 80.44\%, 92.794\%, 16.11\%, and 77.90\% superior forecasting efficacy in winter as compared to the benchmark models.  The proposed model shows 17.49\% greater accuracy when applied to complex terrains as compared to its performance for plain terrains. 

\begin{table}[t!]

\centering
\caption{Season-specific forecasting performance}
\resizebox{1.0\columnwidth}{!}{%
\begin{tabular}{|c|c|c|c|c|c|c|c|}
\hline
\shortstack{Wind\\stations}& \shortstack{Seasons}&\shortstack{LSTM \\\cite{elsaraiti2021comparative}}& \shortstack{EEMD-LSTM \\\cite{huang2018wind}}& \shortstack{EEMD-(BiLSTM)\\\cite{saxena2021offshore} }& \shortstack{(1D-CNN)-LSTM\\\cite{shen2022wind}}&{Proposed approach}\\ 
\hline

    \multicolumn{7}{|c|}{Plain terrain} \\ 
\hline
&Winter&0.1751&0.0461&0.0464&0.1807&\bf0.0669\\
Kayathar&Summer&0.2367&0.0593&0.0586&0.2170&\bf0.0419\\
&Rainy&0.3292&0.0791&0.0773&0.2624&\bf0.0485\\

\hline
&Winter&0.2026&0.1752&0.0550&0.0529&\bf0.0517\\
Kavalkinaru&Summer&0.3486&0.2997&0.0802&0.2869&\bf0.0722\\
&Rainy&0.4684&0.4044&0.1076&0.3701&\bf0.0847\\

\hline
&Winter&0.2539&0.1715&0.1560&0.2266&\bf0.0606\\
Motibaru&Summer&0.3148&0.0930&0.0822&0.3103&\bf0.0563\\
&Rainy&0.5534&0.1042&0.0985&0.5116&\bf0.1095\\

\hline

    \multicolumn{7}{|c|}{Complex terrain} \\ 
\hline
&Winter&0.2570&0.0619&0.0585&0.2161&\bf0.0484\\
Jagmin&Summer&0.2705&0.2663&0.0631&0.2366&\bf0.0449\\
&Rainy&0.3189&0.1309&0.0909&0.2843&\bf0.0686\\

\hline
&Winter&0.2184&0.0672&0.0521&0.2033&\bf0.0443\\
Gadhpati&Summer&0.2768&0.0828&0.0652&0.2745&\bf0.0561\\
&Rainy&0.4412&0.1061&0.0770&0.3293&\bf0.0666\\

\hline

\end{tabular}%
}
\label{tab:my_label80}
\end{table} \par
\begin{figure}[!thb]

\centering

\includegraphics[width=85mm]{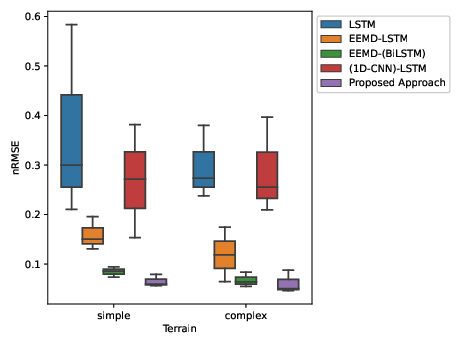}

\caption{Season-specific variability in nRMSE for different terrain conditions}
\label{saeson}
\end{figure}

Figure \ref{saeson}, displays the variability of nRMSE for all models for two types of terrain across three seasons. When dealing with simple terrain conditions, it has been found that LSTM has produced the highest season-wise variability in the nRMSE scores. In contrast, (1D-CNN)-LSTM has produced the maximum seasonal variability in nRMSE scores in the context of complicated terrain. The proposed model demonstrated the lowest mean in complex terrain and the lowest variance in the nRMSE scores for simple terrain conditions, respectively. The divergence of the mean nRMSE scores obtained for the two distinct terrain situations was also computed. The suggested model's deviation was found to be 0.37\%, whereas the LSTM model had a deviation of 6.76\%. The aforementioned analysis, therefore, suggests that the suggested model has demonstrated the most reliable and consistent forecasting performance in both terrain conditions.

\subsection{Overall forecasting performance}

 Table \ref{tab:my_label7} compares the overall forecasting performance of the proposed model to benchmark models in terms of overall nRMSE and mean rank. The proposed approach has shown the lowest forecasting error (in terms of nRMSE) and mean rank as compared to persistence for all forecasting horizons. VMD-(Bi-LSTM) \cite{shen2022wind} is the second-best performing model. 

\begin{table}[t!]

\centering
\caption{Overall forecasting performance of the proposed model compared against benchmarks }
\resizebox{1.0\columnwidth}{!}{%
\begin{tabular}{|c|cc|cc|cc|cc|}
\hline

   \textbf{Models}  &\multicolumn{2}{c}{1 hr}&\multicolumn{2}{c}{2 hrs}&\multicolumn{2}{c}{3 hrs}&\multicolumn{2}{c|}{4 hrs} \\ 
\hline 
  & \textbf{nRMSE}&\textbf{\shortstack{Mean \\rank}}& \textbf{nRMSE}&\textbf{\shortstack{Mean \\rank}}& \textbf{nRMSE}&\textbf{\shortstack{Mean\\ rank}}& \textbf{nRMSE}&\textbf{\shortstack{Mean \\rank}} \\ 
\hline
Persistence &0.2214&7.00&0.3570&7.00&0.6312&7.00&0.6471&7.00\\
LSTM \cite{elsaraiti2021comparative}&0.1800&5.00&0.2751&5.00&0.4377&6.00&0.5271&6.00\\
EEMD-LSTM \cite{huang2018wind}&0.0641	& 3.00&0.0912&3.00&0.1201&3.00&0.1521&4.00\\
EEMD-(Bi-LSTM) \cite{saxena2021offshore}&	0.0569	&  2.00&0.0827&2.00&0.1041&2.00&0.1394&3.00\\
(1D-CNN)-LSTM \cite{shen2022wind}&0.1863&  6.00      &0.3045&6.00&0.4031&5.00 &0.4813&5.00\\
VMD-(Bi-LSTM) \cite{li2022novel}&0.0709&  4.00      &0.1005&4.00&0.1204&4.00 &0.1275&2.00\\
\bf Proposed approach&\bf0.0453&\bf1.00 &\bf0.0685&\bf1.00&\bf0.0806&\bf1.00&\bf0.1136&\bf1.00\\
\hline

\end{tabular}%
}

    \label{tab:my_label7}
\end{table}





\begin{figure}[!htb]
\centering

\begin{tabular}{c}
\includegraphics[width=8cm,height=2.5cm]{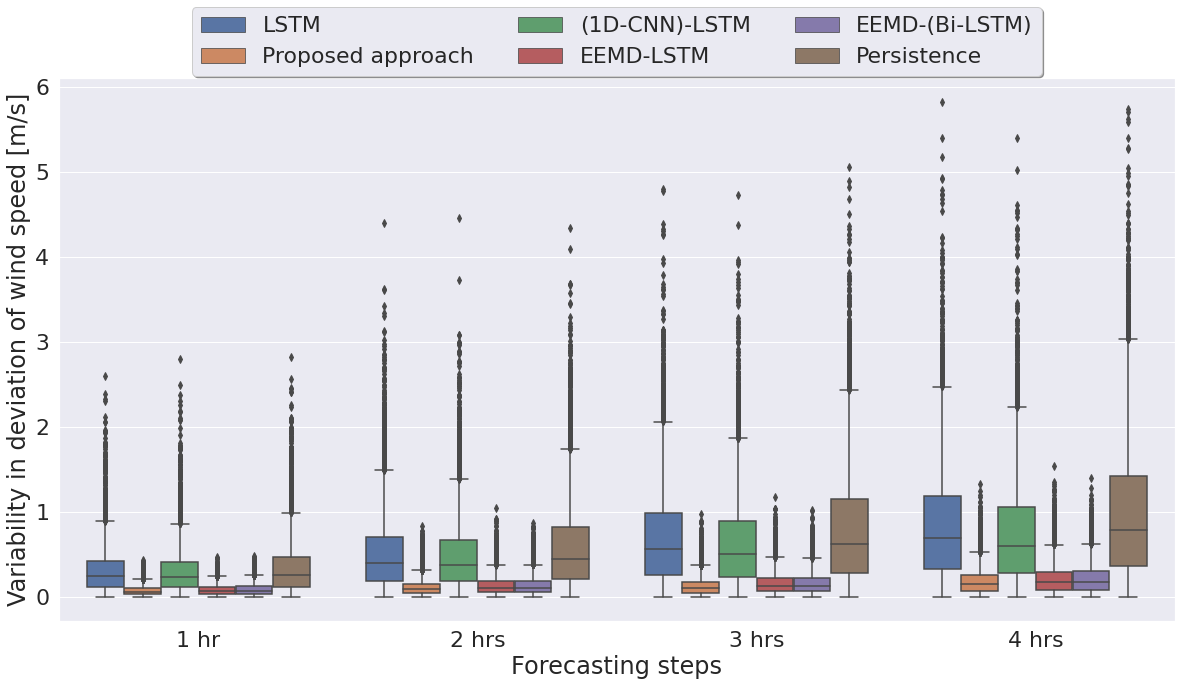}
  \\
  (a) Kayathar \\[6pt]
 \includegraphics[width=8cm,height=2.5cm]{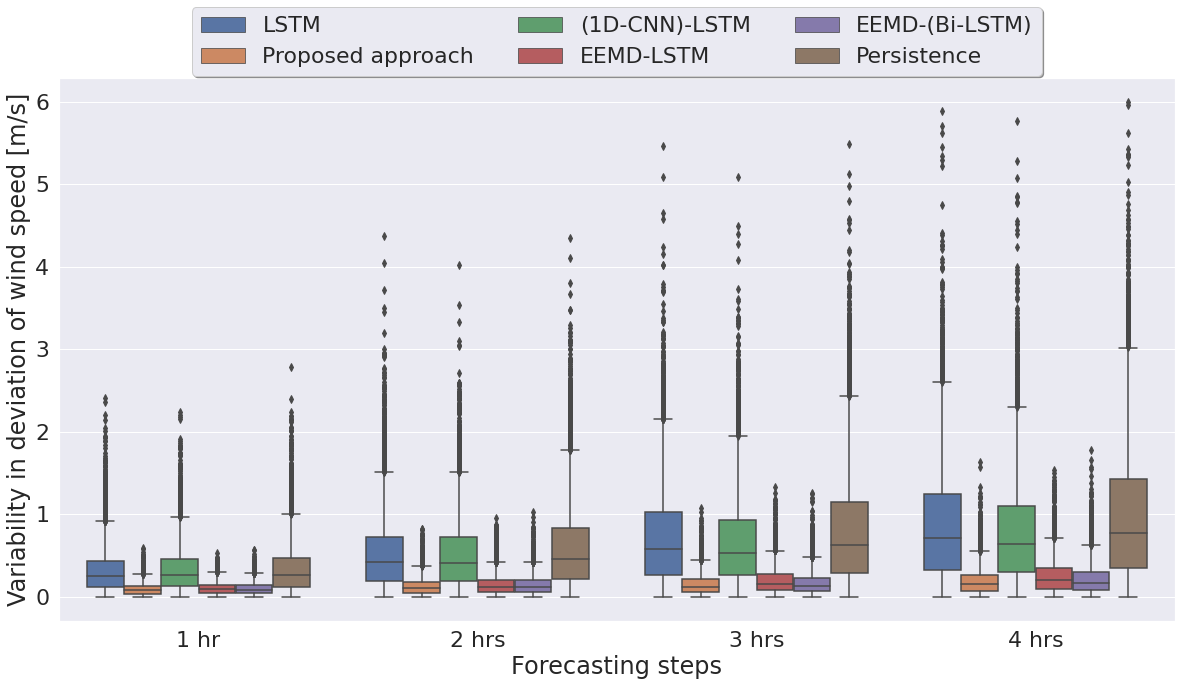}
  \\
    (b) Kavalkinaru\\[6pt]
     \includegraphics[width=8cm,height=2.5cm]{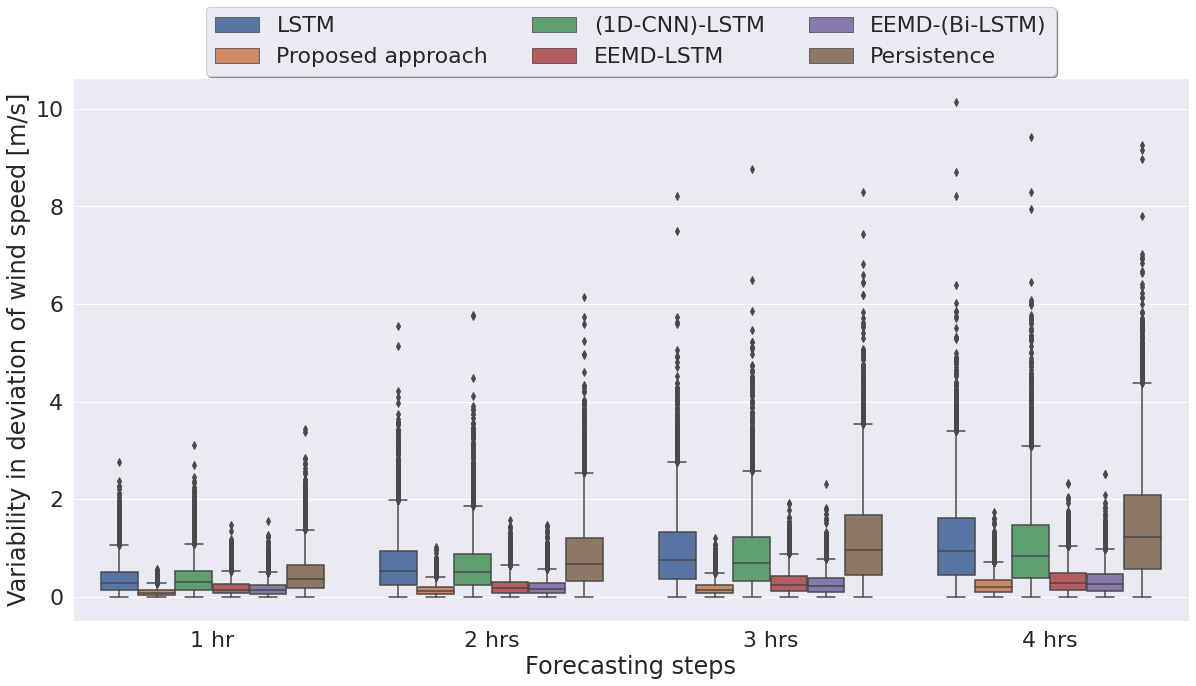}
  \\
  (c) Motibaru \\[6pt]

 \includegraphics[width=8cm,height=2.5cm]{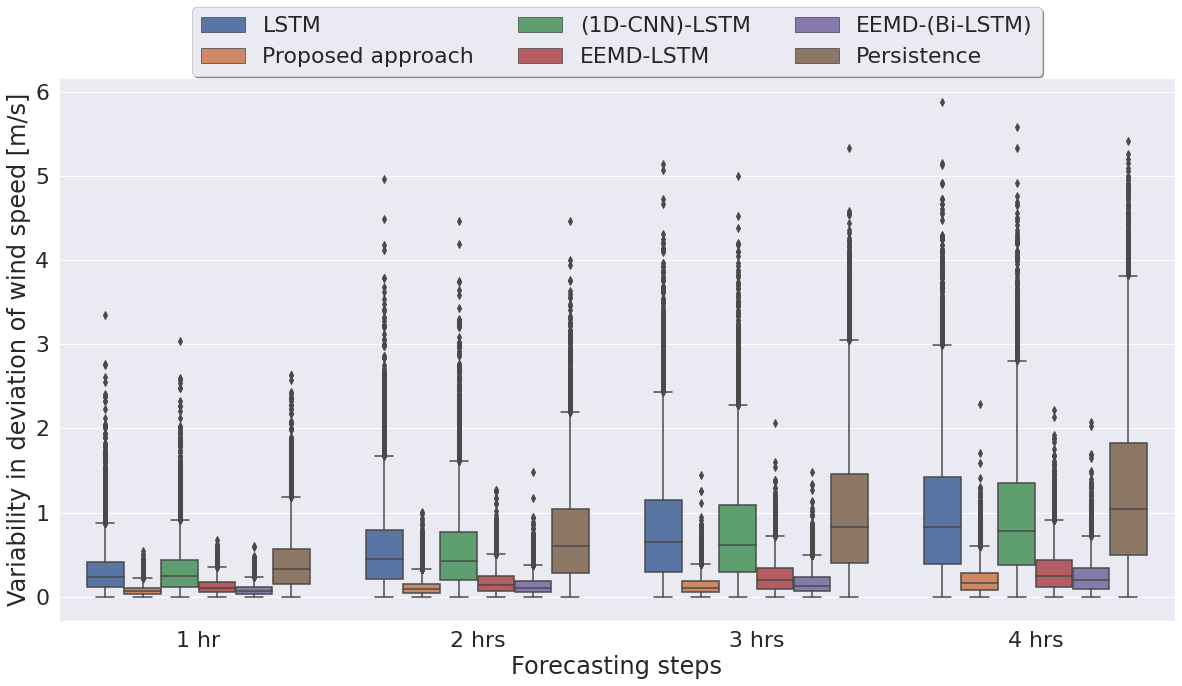}

  \\
  (d) Gadhpati\\[6pt]

    \includegraphics[width=8cm,height=2.5cm]{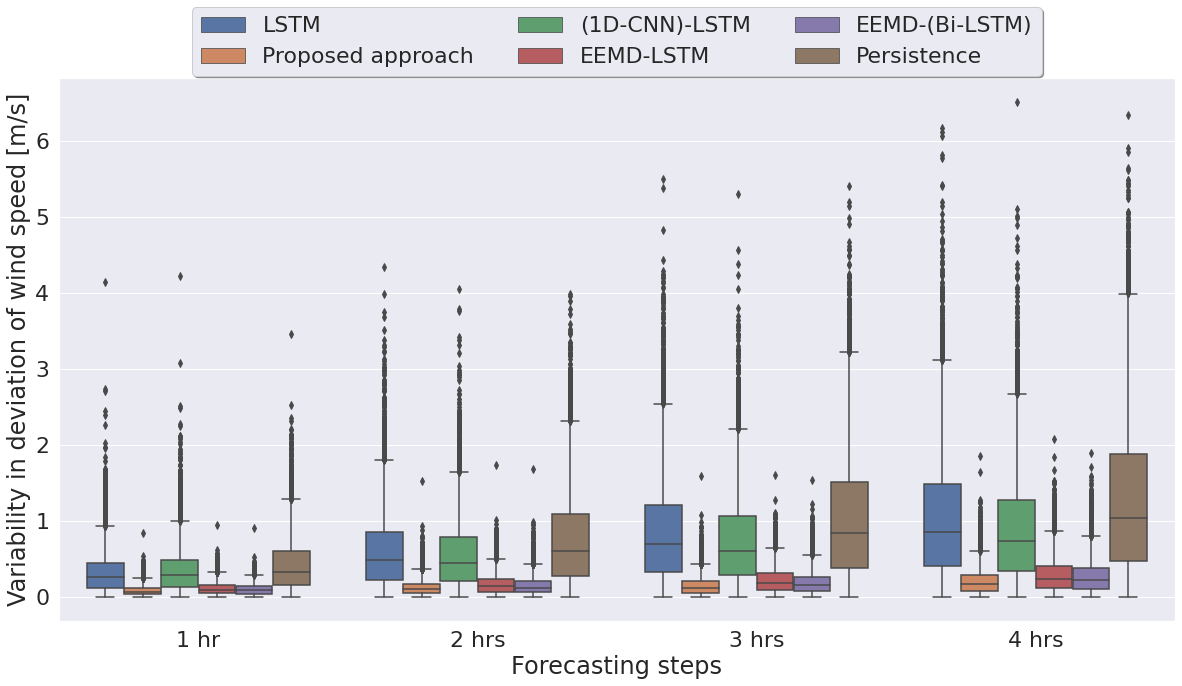}
  \\
  (e) Jagmin \\[6pt]
\end{tabular}
\caption{Variability in the deviation of the WS[m/s] from the actual}
\label{daviation}

\end{figure}

Figure \ref{daviation} shows the variability in the deviation of forecasted WS[m/s] from the actual. Irrespective of the forecasting horizon, the proposed model has achieved the lowest variability in predictions. Also, it has produced the least number of outliers and the lowest mean deviation in its predictions. 

Similarly, a comparison of terrain-wise variability in predictions of the proposed model in terms of nRMSE shows that the proposed model consistently achieves the lowest variability in prediction for all forecasting horizons.  







\section{Conclusion}
\label{con}

A novel adaptive short-term WS forecasting model has been proposed for plain and complex terrain conditions. At first, EEMD was used to decompose the raw WS data into multiple sub-signals. We use PACF to obtain optimal lag for each sub-signal and superimpose sub-signals with similar lags. SampEn is used to calculate the complexity of newly computed sub-signals. Finally, specific deep learning model-feature combinations have been chosen based on the complexity of each sub-signal. The efficiency of the suggested model has been validated for two types of terrain, three seasons, and four forecasting horizons. In terms of RMSE, MAE, and nRMSE, the suggested model is superior to the persistence model, two non-EEMD-based methods \cite{elsaraiti2021comparative, shen2022wind}, and two EEMD-based approaches \cite{huang2018wind,saxena2021offshore}.

 \begin{itemize}
    \item \textbf{Forecasting performance compared to the benchmark:}
     In the case of plain terrain, the proposed model has shown 55.29\%, 48.26\%, and 48.72\% greater forecasting accuracy compared to the benchmark in terms of nRMSE, RMSE, and MAE. In complex terrains, the proposed model has achieved 61.87\%, 56.52\%, and  54.61\% better nRMSE, RMSE, and MAE as compared to the benchmarks. The proposed model has achieved the
lowest variance in terms of forecasting accuracy between simple and complex terrain conditions at 1.75\%.
    \item \textbf{Multi-steps ahead forecasting performance:}
    The proposed model has dominated benchmarks by 54.14\%, 54.34\%, 59.77\%, and 55.51\% in terms of nRMSE for 1 hr, 2 hrs, 3 hrs, and 4 hrs ahead forecasting.
    \item \textbf{Season specific forecasting performance:}
    The proposed model has shown 38.65\%, 57.97\%, and 53.75\% more accurate forecasting performance in winter, summer, and rainy seasons as compared to the benchmarks. In the case of complex terrain, the improvement is 66.81\%, 60.23\%, and 55.23\% in all seasons. The suggested model's mean nRMSE scores for the two different terrain conditions were found to differ by 0.37\% on average.
\end{itemize}

Thus, the performance of the proposed model is found to be robust across seasons and terrain conditions. A future goal is to validate our model across other Indian terrain types. An additional goal is to extend our model to accommodate offshore wind power stations.

\bibliographystyle{unsrtnat}
\bibliography{template}  

\begin{thebibliography}{69}
\providecommand{\natexlab}[1]{#1}
\providecommand{\url}[1]{\texttt{#1}}
\expandafter\ifx\csname urlstyle\endcsname\relax
  \providecommand{\doi}[1]{doi: #1}\else
  \providecommand{\doi}{doi: \begingroup \urlstyle{rm}\Url}\fi

\bibitem[Saidur et~al.(2011)Saidur, Rahim, Islam, and Solangi]{saidur2011environmental}
Rahman Saidur, Nasrudin~A Rahim, Monirul~Rafiqu Islam, and Khalid~H Solangi.
\newblock Environmental impact of wind energy.
\newblock \emph{Renewable and sustainable energy reviews}, 15\penalty0 (5):\penalty0 2423--2430, 2011.
\newblock \url{https://doi.org/10.1016/j.rser.2011.02.024}.

\bibitem[Ahmadi and Khashei(2021)]{ahmadi2021current}
Mehrnaz Ahmadi and Mehdi Khashei.
\newblock Current status of hybrid structures in wind forecasting.
\newblock \emph{Engineering applications of artificial intelligence}, 99:\penalty0 104133, 2021.
\newblock \url{https://doi.org/10.1016/j.engappai.2020.104133}.

\bibitem[Amaral and Castro(2017)]{amaral2017offshore}
Lu{\'\i}s Amaral and Rui Castro.
\newblock Offshore wind farm layout optimization regarding wake effects and electrical losses.
\newblock \emph{Engineering Applications of Artificial Intelligence}, 60:\penalty0 26--34, 2017.
\newblock \url{https://doi.org/10.1016/j.engappai.2017.01.010}.

\bibitem[Khan et~al.(2020)Khan, Kumar, and Richa]{khan2020present}
Kalay Khan, Rohitashw Kumar, and Rishi Richa.
\newblock Present status and future opportunity in renewable energy-a review.
\newblock \emph{SKUAST Journal of Research}, 22\penalty0 (1):\penalty0 19--33, 2020.

\bibitem[Shi et~al.(2013)Shi, Ding, Lee, Yang, Liu, and Zhang]{shi2013hybrid}
Jie Shi, Zhaohao Ding, Wei-Jen Lee, Yongping Yang, Yongqian Liu, and Mingming Zhang.
\newblock Hybrid forecasting model for very-short term wind power forecasting based on grey relational analysis and wind speed distribution features.
\newblock \emph{IEEE Transactions on Smart Grid}, 5\penalty0 (1):\penalty0 521--526, 2013.
\newblock \doi{10.1109/TSG.2013.2283269}.

\bibitem[Viviescas et~al.(2019)Viviescas, Lima, Diuana, Vasquez, Ludovique, Silva, Huback, Magalar, Szklo, Lucena, et~al.]{viviescas2019contribution}
Cindy Viviescas, Lucas Lima, Fabio~A Diuana, Eveline Vasquez, Camila Ludovique, Gabriela~N Silva, Vanessa Huback, Leticia Magalar, Alexandre Szklo, Andre~FP Lucena, et~al.
\newblock Contribution of variable renewable energy to increase energy security in latin america: Complementarity and climate change impacts on wind and solar resources.
\newblock \emph{Renewable and sustainable energy reviews}, 113:\penalty0 109232, 2019.
\newblock \url{https://doi.org/10.1016/j.rser.2019.06.039}.

\bibitem[Abedinia et~al.(2020)Abedinia, Lotfi, Bagheri, Sobhani, Shafie-Khah, and Catal{\~a}o]{abedinia2020improved}
Oveis Abedinia, Mohamed Lotfi, Mehdi Bagheri, Behrouz Sobhani, Miadreza Shafie-Khah, and Jo{\~a}o~PS Catal{\~a}o.
\newblock Improved emd-based complex prediction model for wind power forecasting.
\newblock \emph{IEEE Transactions on Sustainable Energy}, 11\penalty0 (4):\penalty0 2790--2802, 2020.
\newblock \url{https://doi.org/10.1109/TSTE.2020.2976038}.

\bibitem[Tian(2020)]{tian2020short}
Zhongda Tian.
\newblock Short-term wind speed prediction based on lmd and improved fa optimized combined kernel function lssvm.
\newblock \emph{Engineering Applications of Artificial Intelligence}, 91:\penalty0 103573, 2020.
\newblock \url{https://doi.org/10.1016/j.engappai.2020.103573}.

\bibitem[Quan et~al.(2019)Quan, Khosravi, Yang, and Srinivasan]{quan2019survey}
Hao Quan, Abbas Khosravi, Dazhi Yang, and Dipti Srinivasan.
\newblock A survey of computational intelligence techniques for wind power uncertainty quantification in smart grids.
\newblock \emph{IEEE transactions on neural networks and learning systems}, 31\penalty0 (11):\penalty0 4582--4599, 2019.
\newblock \doi{10.1109/TNNLS.2019.2956195}.

\bibitem[Zheng and Zhang(2023)]{zheng2023stochastic}
Zhong Zheng and Zijun Zhang.
\newblock A stochastic recurrent encoder decoder network for multistep probabilistic wind power predictions.
\newblock \emph{IEEE Transactions on Neural Networks and Learning Systems}, 2023.
\newblock \doi{10.1109/TNNLS.2023.3234130}.

\bibitem[Liang and Tang(2022)]{liang2022ultra}
Junkai Liang and Wenyuan Tang.
\newblock Ultra-short-term spatiotemporal forecasting of renewable resources: An attention temporal convolutional network-based approach.
\newblock \emph{IEEE Transactions on Smart Grid}, 13\penalty0 (5):\penalty0 3798--3812, 2022.
\newblock \doi{10.1109/TSG.2022.3175451}.

\bibitem[Wang et~al.(2018)Wang, Hu, Srinivasan, and Wang]{Wang2018Wind}
Yun Wang, Qinghua Hu, Dipti Srinivasan, and Zheng Wang.
\newblock Wind power curve modeling and wind power forecasting with inconsistent data.
\newblock \emph{IEEE Transactions on Sustainable Energy}, 10\penalty0 (1):\penalty0 16--25, 2018.
\newblock \url{https://doi.org/10.1109/TSTE.2018.2820198}.

\bibitem[Taylor et~al.(2009)Taylor, McSharry, and Buizza]{5224014}
James~W. Taylor, Patrick~E. McSharry, and Roberto Buizza.
\newblock Wind power density forecasting using ensemble predictions and time series models.
\newblock \emph{IEEE Transactions on Energy Conversion}, 24\penalty0 (3):\penalty0 775--782, 2009.
\newblock \doi{10.1109/TEC.2009.2025431}.

\bibitem[Arora et~al.(2022)Arora, Jalali, Ahmadian, Panigrahi, Suganthan, and Khosravi]{arora2022probabilistic}
Parul Arora, Seyed Mohammad~Jafar Jalali, Sajad Ahmadian, BK~Panigrahi, PN~Suganthan, and Abbas Khosravi.
\newblock Probabilistic wind power forecasting using optimized deep auto-regressive recurrent neural networks.
\newblock \emph{IEEE Transactions on Industrial Informatics}, 19\penalty0 (3):\penalty0 2814--2825, 2022.
\newblock \url{https://doi.org/10.1109/TII.2022.3160696}.

\bibitem[Yan et~al.(2021)Yan, Hu, Zhen, Wang, Qiu, Li, Yao, Shafie-khah, and Catal{\~a}o]{yan2021frequency}
Jichuan Yan, Lin Hu, Zhao Zhen, Fei Wang, Gang Qiu, Yu~Li, Liangzhong Yao, Miadreza Shafie-khah, and Jo{\~a}o~PS Catal{\~a}o.
\newblock Frequency-domain decomposition and deep learning based solar pv power ultra-short-term forecasting model.
\newblock \emph{IEEE Transactions on Industry Applications}, 57\penalty0 (4):\penalty0 3282--3295, 2021.
\newblock \url{https://doi.org/10.1109/TIA.2021.3073652}.

\bibitem[Yan and Ouyang(2019)]{yan2019advanced}
Jing Yan and Tinghui Ouyang.
\newblock Advanced wind power prediction based on data-driven error correction.
\newblock \emph{Energy conversion and management}, 180:\penalty0 302--311, 2019.
\newblock \url{https://doi.org/10.1016/j.enconman.2018.10.108}.

\bibitem[Wang et~al.(2021)Wang, Zou, Liu, Zhang, and Liu]{wang2021review}
Yun Wang, Runmin Zou, Fang Liu, Lingjun Zhang, and Qianyi Liu.
\newblock A review of wind speed and wind power forecasting with deep neural networks.
\newblock \emph{Applied Energy}, 304:\penalty0 117766, 2021.
\newblock \url{https://doi.org/10.1016/j.apenergy.2021.117766}.

\bibitem[Sharples et~al.(2010)Sharples, McRae, and Weber]{sharples2010wind}
Jason~J Sharples, Richard~HD McRae, and RO~Weber.
\newblock Wind characteristics over complex terrain with implications for bushfire risk management.
\newblock \emph{Environmental Modelling \& Software}, 25\penalty0 (10):\penalty0 1099--1120, 2010.
\newblock \url{https://doi.org/10.1016/j.envsoft.2010.03.016}.

\bibitem[Tian et~al.(2015)Tian, Ozbay, and Hu]{tian2015terrain}
Wei Tian, Ahmet Ozbay, and Hui Hu.
\newblock Terrain effects on characteristics of surface wind and wind turbine wakes.
\newblock \emph{Procedia Engineering}, 126:\penalty0 542--548, 2015.
\newblock \url{https://doi.org/10.1016/j.proeng.2015.11.302}.

\bibitem[Farrell et~al.(2021)Farrell, King, Draxl, Mudafort, Hamilton, Bay, Fleming, and Simley]{farrell2021design}
Alayna Farrell, Jennifer King, Caroline Draxl, Rafael Mudafort, Nicholas Hamilton, Christopher~J Bay, Paul Fleming, and Eric Simley.
\newblock Design and analysis of a wake model for spatially heterogeneous flow.
\newblock \emph{Wind Energy Science}, 6\penalty0 (3):\penalty0 737--758, 2021.
\newblock \url{https://doi.org/10.5194/wes-6-737-2021}.

\bibitem[Huang et~al.(2018)Huang, Liu, and Yang]{huang2018wind}
Yuansheng Huang, Shijian Liu, and Lei Yang.
\newblock Wind speed forecasting method using eemd and the combination forecasting method based on gpr and lstm.
\newblock \emph{Sustainability}, 10\penalty0 (10):\penalty0 3693, 2018.
\newblock \url{ https://doi.org/10.3390/su10103693}.

\bibitem[Shen et~al.(2022)Shen, Fan, Zhang, and Yu]{shen2022wind}
Zhipeng Shen, Xuechun Fan, Liangyu Zhang, and Haomiao Yu.
\newblock Wind speed prediction of unmanned sailboat based on cnn and lstm hybrid neural network.
\newblock \emph{Ocean Engineering}, 254:\penalty0 111352, 2022.
\newblock \url{https://doi.org/10.1016/j.oceaneng.2022.111352}.

\bibitem[Saxena et~al.(2021{\natexlab{a}})Saxena, Mishra, and Rao]{saxena2021offshore}
Bharat~Kumar Saxena, Sanjeev Mishra, and Komaragiri Venkata~Subba Rao.
\newblock Offshore wind speed forecasting at different heights by using ensemble empirical mode decomposition and deep learning models.
\newblock \emph{Applied Ocean Research}, 117:\penalty0 102937, 2021{\natexlab{a}}.
\newblock \url{https://doi.org/10.1016/j.apor.2021.102937}.

\bibitem[Zhou et~al.(2011)Zhou, Shi, and Li]{zhou2011fine}
Junyi Zhou, Jing Shi, and Gong Li.
\newblock Fine tuning support vector machines for short-term wind speed forecasting.
\newblock \emph{Energy Conversion and Management}, 52\penalty0 (4):\penalty0 1990--1998, 2011.
\newblock \url{https://doi.org/10.1016/j.enconman.2010.11.007}.

\bibitem[Zuluaga et~al.(2015)Zuluaga, Alvarez, and Giraldo]{zuluaga2015short}
Carlos~D Zuluaga, Mauricio~A Alvarez, and Eduardo Giraldo.
\newblock Short-term wind speed prediction based on robust kalman filtering: An experimental comparison.
\newblock \emph{Applied Energy}, 156:\penalty0 321--330, 2015.
\newblock \url{https://doi.org/10.1016/j.apenergy.2015.07.043}.

\bibitem[Zhang et~al.(2022)Zhang, Peng, and Nazir]{zhang2022novel}
Chu Zhang, Tian Peng, and Muhammad~Shahzad Nazir.
\newblock A novel hybrid approach based on variational heteroscedastic gaussian process regression for multi-step ahead wind speed forecasting.
\newblock \emph{International Journal of Electrical Power \& Energy Systems}, 136:\penalty0 107717, 2022.
\newblock \url{https://doi.org/10.1016/j.ijepes.2021.107717}.

\bibitem[Hua et~al.(2022)Hua, Zhang, Peng, Ji, and Nazir]{hua2022integrated}
Lei Hua, Chu Zhang, Tian Peng, Chunlei Ji, and Muhammad~Shahzad Nazir.
\newblock Integrated framework of extreme learning machine (elm) based on improved atom search optimization for short-term wind speed prediction.
\newblock \emph{Energy Conversion and Management}, 252:\penalty0 115102, 2022.
\newblock \url{https://doi.org/10.1016/j.enconman.2021.115102}.

\bibitem[Samadianfard et~al.(2020)Samadianfard, Hashemi, Kargar, Izadyar, Mostafaeipour, Mosavi, Nabipour, and Shamshirband]{samadianfard2020wind}
Saeed Samadianfard, Sajjad Hashemi, Katayoun Kargar, Mojtaba Izadyar, Ali Mostafaeipour, Amir Mosavi, Narjes Nabipour, and Shahaboddin Shamshirband.
\newblock Wind speed prediction using a hybrid model of the multi-layer perceptron and whale optimization algorithm.
\newblock \emph{Energy Reports}, 6:\penalty0 1147--1159, 2020.
\newblock \url{https://doi.org/10.1016/j.egyr.2020.05.001}.

\bibitem[Navas et~al.(2020)Navas, Prakash, and Sasipraba]{navas2020artificial}
R~Kaja~Bantha Navas, S~Prakash, and T~Sasipraba.
\newblock Artificial neural network based computing model for wind speed prediction: A case study of coimbatore, tamil nadu, india.
\newblock \emph{Physica A: Statistical Mechanics and its Applications}, 542:\penalty0 123383, 2020.
\newblock \url{https://doi.org/10.1016/j.physa.2019.123383}.

\bibitem[Lv and Yue(2011)]{lv2011short}
Peng Lv and Lili Yue.
\newblock Short-term wind speed forecasting based on non-stationary time series analysis and arch model.
\newblock In \emph{2011 International Conference on Multimedia Technology}, pages 2549--2553. IEEE, 2011.

\bibitem[Li et~al.(2022{\natexlab{a}})Li, Wang, Zhang, and Li]{li2022innovative}
Jingrui Li, Jianzhou Wang, Haipeng Zhang, and Zhiwu Li.
\newblock An innovative combined model based on multi-objective optimization approach for forecasting short-term wind speed: A case study in china.
\newblock \emph{Renewable Energy}, 201:\penalty0 766--779, 2022{\natexlab{a}}.
\newblock \url{https://doi.org/10.1016/j.renene.2022.10.123}.

\bibitem[Hu et~al.(2021)Hu, Yang, Chen, Yuan, Li, Shao, and Zhang]{hu2021new}
Weicheng Hu, Qingshan Yang, Hua-Peng Chen, Ziting Yuan, Chen Li, Shuai Shao, and Jian Zhang.
\newblock New hybrid approach for short-term wind speed predictions based on preprocessing algorithm and optimization theory.
\newblock \emph{Renewable Energy}, 179:\penalty0 2174--2186, 2021.
\newblock \url{https://doi.org/10.1016/j.renene.2021.08.044}.

\bibitem[Abdulla et~al.(2022)Abdulla, Demirci, and Ozdemir]{abdulla2022design}
Nawaf Abdulla, Mehmet Demirci, and Suat Ozdemir.
\newblock Design and evaluation of adaptive deep learning models for weather forecasting.
\newblock \emph{Engineering Applications of Artificial Intelligence}, 116:\penalty0 105440, 2022.
\newblock \url{https://doi.org/10.1016/j.engappai.2022.105440}.

\bibitem[Memarzadeh and Keynia(2020)]{memarzadeh2020new}
Gholamreza Memarzadeh and Farshid Keynia.
\newblock A new short-term wind speed forecasting method based on fine-tuned lstm neural network and optimal input sets.
\newblock \emph{Energy Conversion and Management}, 213:\penalty0 112824, 2020.
\newblock \url{https://doi.org/10.1016/j.enconman.2020.112824}.

\bibitem[Wang et~al.(2020)Wang, Zhang, Zhang, Huang, and Wang]{wang2020short}
Zhongju Wang, Jing Zhang, Yu~Zhang, Chao Huang, and Long Wang.
\newblock Short-term wind speed forecasting based on information of neighboring wind farms.
\newblock \emph{IEEE Access}, 8:\penalty0 16760--16770, 2020.
\newblock \url{https://doi.org/10.1109/ACCESS.2020.2966268}.

\bibitem[Saxena et~al.(2021{\natexlab{b}})Saxena, Mishra, and Rao]{saxena2021application}
Bharat~Kumar Saxena, Sanjeev Mishra, and Komaragiri Venkata~Subba Rao.
\newblock Application of stacked and bidirectional long short-term memory deep learning models for wind speed forecasting at an offshore site.
\newblock \emph{Energy Sources, Part A: Recovery, Utilization, and Environmental Effects}, pages 1--16, 2021{\natexlab{b}}.
\newblock \url{https://doi.org/10.1080/15567036.2021.1925379}.

\bibitem[Syu et~al.(2020)Syu, Wang, Chou, Lin, Liang, Wu, and Jiang]{syu2020ultra}
Yu-Dian Syu, Jen-Cheng Wang, Cheng-Ying Chou, Ming-Jhou Lin, Wei-Chih Liang, Li-Cheng Wu, and Joe-Air Jiang.
\newblock Ultra-short-term wind speed forecasting for wind power based on gated recurrent unit.
\newblock In \emph{2020 8th International Electrical Engineering Congress (iEECON)}, pages 1--4. IEEE, 2020.
\newblock \url{https://doi.org/10.1109/iEECON48109.2020.229518}.

\bibitem[Xu et~al.(2022)Xu, Yang, Luo, He, and Sun]{xu2022multi}
Yuanyuan Xu, Genke Yang, Jiliang Luo, Jianan He, and Haixin Sun.
\newblock A multi-location short-term wind speed prediction model based on spatiotemporal joint learning.
\newblock \emph{Renewable Energy}, 183:\penalty0 148--159, 2022.
\newblock \url{https://doi.org/10.1016/j.renene.2021.10.075}.

\bibitem[Hassani(2007)]{hassani2007singular}
Hossein Hassani.
\newblock Singular spectrum analysis: methodology and comparison.
\newblock 2007.

\bibitem[Dragomiretskiy and Zosso(2013)]{dragomiretskiy2013variational}
Konstantin Dragomiretskiy and Dominique Zosso.
\newblock Variational mode decomposition.
\newblock \emph{IEEE transactions on signal processing}, 62\penalty0 (3):\penalty0 531--544, 2013.
\newblock \url{https://doi.org/10.1109/TSP.2013.2288675}.

\bibitem[Wan et~al.(2021)Wan, Qian, Zhao, Song, and Yang]{9384295}
Can Wan, Weiting Qian, Changfei Zhao, Yonghua Song, and Guangya Yang.
\newblock Probabilistic forecasting based sizing and control of hybrid energy storage for wind power smoothing.
\newblock \emph{IEEE Transactions on Sustainable Energy}, 12\penalty0 (4):\penalty0 1841--1852, 2021.
\newblock \doi{10.1109/TSTE.2021.3068043}.

\bibitem[Li et~al.(2019)Li, Tang, Xue, Saeed, and Hu]{li2019short}
Chaoshun Li, Geng Tang, Xiaoming Xue, Adnan Saeed, and Xin Hu.
\newblock Short-term wind speed interval prediction based on ensemble gru model.
\newblock \emph{IEEE transactions on sustainable energy}, 11\penalty0 (3):\penalty0 1370--1380, 2019.
\newblock \url{https://doi.org/10.1109/TSTE.2019.2926147}.

\bibitem[Fu et~al.(2021)Fu, Fang, Wang, Li, Xiong, and Zhang]{fu2021multi}
Wenlong Fu, Ping Fang, Kai Wang, Zhenxing Li, Dongzhen Xiong, and Kai Zhang.
\newblock Multi-step ahead short-term wind speed forecasting approach coupling variational mode decomposition, improved beetle antennae search algorithm-based synchronous optimization and volterra series model.
\newblock \emph{Renewable Energy}, 179:\penalty0 1122--1139, 2021.
\newblock \url{https://doi.org/10.1016/j.renene.2021.07.119}.

\bibitem[Wang et~al.(2019)Wang, Li, Fu, and Tang]{wang2019deep}
Ruoheng Wang, Chaoshun Li, Wenlong Fu, and Geng Tang.
\newblock Deep learning method based on gated recurrent unit and variational mode decomposition for short-term wind power interval prediction.
\newblock \emph{IEEE transactions on neural networks and learning systems}, 31\penalty0 (10):\penalty0 3814--3827, 2019.
\newblock \doi{10.1109/TNNLS.2019.2946414}.

\bibitem[Rilling et~al.(2003)Rilling, Flandrin, Goncalves, et~al.]{rilling2003empirical}
Gabriel Rilling, Patrick Flandrin, Paulo Goncalves, et~al.
\newblock On empirical mode decomposition and its algorithms.
\newblock In \emph{IEEE-EURASIP workshop on nonlinear signal and image processing}, volume~3, pages 8--11. Citeseer, 2003.

\bibitem[Ren et~al.(2014)Ren, Suganthan, and Srikanth]{ren2014novel}
Ye~Ren, Ponnuthurai~Nagaratnam Suganthan, and Narasimalu Srikanth.
\newblock A novel empirical mode decomposition with support vector regression for wind speed forecasting.
\newblock \emph{IEEE transactions on neural networks and learning systems}, 27\penalty0 (8):\penalty0 1793--1798, 2014.
\newblock \doi{10.1109/TNNLS.2014.2351391}.

\bibitem[Wu and Huang(2009)]{wu2009ensemble}
Zhaohua Wu and Norden~E Huang.
\newblock Ensemble empirical mode decomposition: a noise-assisted data analysis method.
\newblock \emph{Advances in adaptive data analysis}, 1\penalty0 (01):\penalty0 1--41, 2009.
\newblock \url{https://doi.org/10.1142/S1793536909000047}.

\bibitem[Manjula and Sarma(2012)]{manjula2012comparison}
M~Manjula and AVRS Sarma.
\newblock Comparison of empirical mode decomposition and wavelet based classification of power quality events.
\newblock \emph{Energy Procedia}, 14:\penalty0 1156--1162, 2012.
\newblock \url{https://doi.org/10.1016/j.egypro.2011.12.1069}.

\bibitem[Li et~al.(2022{\natexlab{b}})Li, Song, Wang, Wang, and Jia]{li2022novel}
Jiale Li, Zihao Song, Xuefei Wang, Yanru Wang, and Yaya Jia.
\newblock A novel offshore wind farm typhoon wind speed prediction model based on pso--bi-lstm improved by vmd.
\newblock \emph{Energy}, 251:\penalty0 123848, 2022{\natexlab{b}}.
\newblock \url{https://doi.org/10.1016/j.energy.2022.123848}.

\bibitem[Chen et~al.(2021)Chen, Dong, Wang, Su, Han, Zhou, Zhang, Zhao, and Bao]{chen2021short}
Yaoran Chen, Zhikun Dong, Yan Wang, Jie Su, Zhaolong Han, Dai Zhou, Kai Zhang, Yongsheng Zhao, and Yan Bao.
\newblock Short-term wind speed predicting framework based on eemd-ga-lstm method under large scaled wind history.
\newblock \emph{Energy Conversion and Management}, 227:\penalty0 113559, 2021.
\newblock \url{https://doi.org/10.1016/j.enconman.2020.113559}.

\bibitem[Duan et~al.(2022)Duan, Chang, Chen, Wang, Zuo, Bai, and Chen]{duan2022combined}
Jikai Duan, Mingheng Chang, Xiangyue Chen, Wenpeng Wang, Hongchao Zuo, Yulong Bai, and Bolong Chen.
\newblock A combined short-term wind speed forecasting model based on cnn--rnn and linear regression optimization considering error.
\newblock \emph{Renewable Energy}, 200:\penalty0 788--808, 2022.
\newblock \url{https://doi.org/10.1016/j.renene.2022.09.114}.

\bibitem[Lee and Baldick(2013)]{lee2013short}
Duehee Lee and Ross Baldick.
\newblock Short-term wind power ensemble prediction based on gaussian processes and neural networks.
\newblock \emph{IEEE Transactions on Smart Grid}, 5\penalty0 (1):\penalty0 501--510, 2013.
\newblock \doi{10.1109/TSG.2013.2280649}.

\bibitem[Luo et~al.(2018)Luo, Sun, Wang, Wang, Zhao, Wu, Wang, and Zhang]{luo2018short}
Xiong Luo, Jiankun Sun, Long Wang, Weiping Wang, Wenbing Zhao, Jinsong Wu, Jenq-Haur Wang, and Zijun Zhang.
\newblock Short-term wind speed forecasting via stacked extreme learning machine with generalized correntropy.
\newblock \emph{IEEE Transactions on Industrial Informatics}, 14\penalty0 (11):\penalty0 4963--4971, 2018.
\newblock \url{https://doi.org/10.1109/TII.2018.2854549}.

\bibitem[Song et~al.(2010)Song, Li{\`o}, et~al.]{song2010new}
Yuedong Song, Pietro Li{\`o}, et~al.
\newblock A new approach for epileptic seizure detection: sample entropy based feature extraction and extreme learning machine.
\newblock \emph{Journal of Biomedical Science and Engineering}, 3\penalty0 (06):\penalty0 556, 2010.

\bibitem[Richman and Moorman(2000)]{richman2000physiological}
Joshua~S Richman and J~Randall Moorman.
\newblock Physiological time-series analysis using approximate entropy and sample entropy.
\newblock \emph{American journal of physiology-heart and circulatory physiology}, 2000.
\newblock \url{https://doi.org/10.1152/ajpheart.2000.278.6.H2039}.

\bibitem[Bagherzadeh and Salehi(2021)]{bagherzadeh2021analysis}
Seyed~Amin Bagherzadeh and Mehdi Salehi.
\newblock Analysis of in-flight cabin vibration of a turboprop airplane by proposing a novel noise-tolerant signal decomposition method.
\newblock \emph{Journal of Vibration and Control}, page 10775463211007583, 2021.
\newblock \url{https://doi.org/10.1177/10775463211007583}.

\bibitem[Gao et~al.(2008)Gao, Ge, Sheng, and Sang]{4566821}
Yunchao Gao, Guangtao Ge, Zhengyan Sheng, and Enfang Sang.
\newblock Analysis and solution to the mode mixing phenomenon in emd.
\newblock In \emph{2008 Congress on Image and Signal Processing}, volume~5, pages 223--227, 2008.
\newblock \doi{10.1109/CISP.2008.193}.

\bibitem[Schmidhuber et~al.(1997)Schmidhuber, Hochreiter, et~al.]{schmidhuber1997long}
J{\"u}rgen Schmidhuber, Sepp Hochreiter, et~al.
\newblock Long short-term memory.
\newblock \emph{Neural Comput}, 9\penalty0 (8):\penalty0 1735--1780, 1997.
\newblock \url{https://doi.org/10.1162/neco.1997.9.8.1735}.

\bibitem[Sharma et~al.(2021)Sharma, Mangla, Yadav, Goyal, Singh, Verma, and Saber]{sharma2021sequential}
Nonita Sharma, Monika Mangla, Sourabh Yadav, Nitin Goyal, Aman Singh, Sahil Verma, and Takfarinas Saber.
\newblock A sequential ensemble model for photovoltaic power forecasting.
\newblock \emph{Computers \& Electrical Engineering}, 96:\penalty0 107484, 2021.
\newblock \url{https://doi.org/10.1016/j.compeleceng.2021.107484}.

\bibitem[Gelaro et~al.(2017)Gelaro, McCarty, Su{\'a}rez, Todling, Molod, Takacs, Randles, Darmenov, Bosilovich, Reichle, et~al.]{gelaro2017modern}
Ronald Gelaro, Will McCarty, Max~J Su{\'a}rez, Ricardo Todling, Andrea Molod, Lawrence Takacs, Cynthia~A Randles, Anton Darmenov, Michael~G Bosilovich, Rolf Reichle, et~al.
\newblock The modern-era retrospective analysis for research and applications, version 2 (merra-2).
\newblock \emph{Journal of climate}, 30\penalty0 (14):\penalty0 5419--5454, 2017.
\newblock \url{https://doi.org/10.1175/JCLI-D-16-0758.1}.

\bibitem[Richman et~al.(2004)Richman, Lake, and Moorman]{richman2004sample}
Joshua~S Richman, Douglas~E Lake, and J~Randall Moorman.
\newblock Sample entropy.
\newblock In \emph{Methods in enzymology}, volume 384, pages 172--184. Elsevier, 2004.
\newblock \url{https://doi.org/10.1016/S0076-6879(04)84011-4}.

\bibitem[Patro and Sahu(2015)]{patro2015normalization}
S~Patro and Kishore~Kumar Sahu.
\newblock Normalization: A preprocessing stage.
\newblock \emph{arXiv preprint arXiv:1503.06462}, 2015.

\bibitem[Chen et~al.(2022)Chen, Yu, Islam, Lim, and Muyeen]{chen2022decomposition}
Yinsong Chen, Samson Yu, Shama Islam, Chee~Peng Lim, and SM~Muyeen.
\newblock Decomposition-based wind power forecasting models and their boundary issue: An in-depth review and comprehensive discussion on potential solutions.
\newblock \emph{Energy Reports}, 8:\penalty0 8805--8820, 2022.
\newblock \url{https://doi.org/10.1016/j.egyr.2022.07.005}.

\bibitem[Gulli and Pal(2017)]{gulli2017deep}
Antonio Gulli and Sujit Pal.
\newblock \emph{Deep learning with Keras}.
\newblock Packt Publishing Ltd, 2017.

\bibitem[Abadi et~al.(2016)Abadi, Barham, Chen, Chen, Davis, Dean, Devin, Ghemawat, Irving, Isard, et~al.]{abadi2016tensorflow}
Mart{\'\i}n Abadi, Paul Barham, Jianmin Chen, Zhifeng Chen, Andy Davis, Jeffrey Dean, Matthieu Devin, Sanjay Ghemawat, Geoffrey Irving, Michael Isard, et~al.
\newblock $\{$TensorFlow$\}$: a system for $\{$Large-Scale$\}$ machine learning.
\newblock In \emph{12th USENIX symposium on operating systems design and implementation (OSDI 16)}, pages 265--283, 2016.

\bibitem[Hunter(2007)]{hunter2007matplotlib}
John~D Hunter.
\newblock Matplotlib: A 2d graphics environment.
\newblock \emph{Computing in science \& engineering}, 9\penalty0 (03):\penalty0 90--95, 2007.
\newblock \url{ https://doi.org/10.1109/MCSE.2007.55}.

\bibitem[Bisong(2019)]{bisong2019matplotlib}
Ekaba Bisong.
\newblock Matplotlib and seaborn.
\newblock In \emph{Building machine learning and deep learning models on google cloud platform}, pages 151--165. Springer, 2019.
\newblock \url{ https://doi.org/10.1007/978-1-4842-4470-8_12}.

\bibitem[Bergstra et~al.(2011)Bergstra, Bardenet, Bengio, and K{\'e}gl]{bergstra2011algorithms}
James Bergstra, R{\'e}mi Bardenet, Yoshua Bengio, and Bal{\'a}zs K{\'e}gl.
\newblock Algorithms for hyper-parameter optimization.
\newblock \emph{Advances in neural information processing systems}, 24, 2011.

\bibitem[Elsaraiti and Merabet(2021)]{elsaraiti2021comparative}
Meftah Elsaraiti and Adel Merabet.
\newblock A comparative analysis of the arima and lstm predictive models and their effectiveness for predicting wind speed.
\newblock \emph{Energies}, 14\penalty0 (20):\penalty0 6782, 2021.
\newblock \url{https://doi.org/10.3390/en14206782}.

\end{thebibliography}






\end{document}